%% file: main.tex
\documentclass[journal,twoside,web]{ieeecolor}

\usepackage{etoolbox}
\makeatletter
\@ifundefined{color@begingroup}%
{\let\color@begingroup\relax
\let\color@endgroup\relax}{}%
\def\fix@ieeecolor@hbox#1{%
\hbox{\color@begingroup#1\color@endgroup}}
\patchcmd\@makecaption{\hbox}{\fix@ieeecolor@hbox}{}{\FAILED}
\patchcmd\@makecaption{\hbox}{\fix@ieeecolor@hbox}{}{\FAILED}

\usepackage{rotating}
\usepackage{tmi}
\usepackage{amsmath,amssymb,amsfonts}
\usepackage{algorithmic}
\usepackage{graphicx}
\usepackage{textcomp}

\DeclareUnicodeCharacter{2212}{-}

\usepackage{multirow}
\usepackage{array}
\usepackage{amssymb}
\usepackage[caption=false,font=normalsize,labelfont=sf,textfont=sf]{subfig}

\usepackage{booktabs}
\usepackage{stfloats}
\usepackage{url}
\usepackage{color}
\usepackage{xspace}
\usepackage[table]{xcolor}
\usepackage{colortbl}
\usepackage{tabularx}
\usepackage{bm}

\makeatletter
\let\NAT@parse\undefined
\makeatother

\usepackage[colorlinks=true, citecolor=blue, linkcolor=blue, urlcolor=blue]{hyperref}
\usepackage{cite}
\usepackage{xargs} 
\usepackage[pdftex,dvipsnames]{xcolor}

\usepackage[colorinlistoftodos,textsize=scriptsize,textwidth=8mm, disable]{todonotes}
\usepackage{marginnote}

\newcommandx{\unsure}[2][1=]{\todo[linecolor=red,backgroundcolor=red!25,bordercolor=red,#1]{#2}}
\newcommandx{\change}[2][1=]{\todo[linecolor=blue,backgroundcolor=blue!25,bordercolor=blue,#1]{#2}}
\newcommandx{\info}[2][1=]{\todo[linecolor=OliveGreen,backgroundcolor=OliveGreen!25,bordercolor=OliveGreen,#1]{#2}}
\newcommandx{\improvement}[2][1=]{\todo[linecolor=Plum,backgroundcolor=Plum!25,bordercolor=Plum,#1]{#2}}
\newcommandx{\thiswillnotshow}[2][1=]{\todo[disable,#1]{#2}}

\newcommand{\infol}[2][]{{%
 \let\marginpar\marginnote
 \reversemarginpar\info[#1]{{\bf #2}}}}

\newcommand{\blue}[1]{{#1}}

\newcommand{\mat}[1]{\mathbf{#1}} % matrix

\newcommand{\etal}{\textit{et al}.\xspace}
\newcommand{\ie}{\textit{i}.\textit{e}.\xspace}
\newcommand{\eg}{\textit{e}.\textit{g}.\xspace}

\newcommand{\modelname}{FoundDiff\xspace}
\newcommand{\stageone}{DA-CLIP\xspace}
\newcommand{\stagetwo}{DA-Diff\xspace}
\newcommand{\block}{DACB\xspace}

\renewcommand{\paragraph}[1]{\noindent\textbf{#1}}

\newlength\savewidth\newcommand\shline{\noalign{\global\savewidth\arrayrulewidth
  \global\arrayrulewidth 1.5pt}\hline\noalign{\global\arrayrulewidth\savewidth}}

\def\BibTeX{{\rm B\kern-.05em{\sc i\kern-.025em b}\kern-.08em
    T\kern-.1667em\lower.7ex\hbox{E}\kern-.125emX}}
\begin{document}
\title{\modelname: Foundational Diffusion Model for Generalizable Low-Dose CT Denoising}

\author{Zhihao~Chen,~Qi~Gao,~Zilong~Li,~Junping Zhang,~\IEEEmembership{Senior~Member,~IEEE}, Yi~Zhang,~\IEEEmembership{Senior~Member,~IEEE},~Jun Zhao,~\IEEEmembership{Member,~IEEE},~and~Hongming~Shan,~\IEEEmembership{Senior~Member,~IEEE}
\thanks{This work was supported in part by the National Natural Science Foundation of China (No. 62471148). (Corresponding author: H. Shan).}
\thanks{Z. Chen, Q. Gao and H. Shan are with the Institute of Science and Technology for Brain-inspired Intelligence and MOE Frontiers Center for Brain Science, Fudan University, Shanghai 200433, China, and also with Shanghai Center for Brain Science and Brain-inspired Technology, Shanghai 201602, China (e-mail: zhihaochen21@m.fudan.edu.cn; qgao21@m.fudan.edu.cn; hmshan@fudan.edu.cn)}
\thanks{Z. Li and J. Zhang are with the Shanghai Key Lab of Intelligent Information Processing, School of Computer Science, Fudan University, Shanghai 200433, China (e-mail: zilongli21@m.fudan.edu.cn; jpzhang@fudan.edu.cn)}
\thanks{Y. Zhang is with School of Cyber Science and Engineering, Sichuan University, Chengdu, Sichuan 610065, China (e-mail: yzhang@scu.edu.cn)}
\thanks{J. Zhao is with the School of Biomedical Engineering, Shanghai Jiao Tong University, Shanghai 200240, China (e-mail: junzhao@sjtu.edu.cn)}
}

\maketitle

\input{sections/abstract.tex}
\input{sections/introduction.tex}
%\input{section/related_work.tex}
\input{sections/method.tex}

\input{sections/experiment.tex}

\input{sections/discussion}
\input{sections/conclusion.tex}

% \bibliographystyle{IEEEtran}
% \bibliography{IEEEabrv,references}

\end{document}

%% file: sections/abstract.tex
\begin{abstract}
% In medical imaging, 
Low-dose computed tomography (CT) denoising is crucial for reduced radiation exposure  while ensuring diagnostically acceptable image quality. 
% Despite significant advancements driven by deep learning (DL) in recent years, achieving generalizability and robustness in low-dose CT (LDCT) denoising remains a persistent challenge. This is primarily because most DL-based methods are trained on specific dose levels and anatomical regions, .
Despite significant advancements driven by deep learning (DL) in recent years, existing DL-based methods, typically trained on a specific dose level and anatomical region, struggle to handle diverse noise characteristics and anatomical heterogeneity during varied scanning conditions, limiting their generalizability and robustness in clinical scenarios.
In this paper, we propose \modelname, a foundational diffusion model for unified and generalizable LDCT denoising across various dose levels and anatomical regions. 
\modelname employs a two-stage strategy: \textbf{(i)} dose-anatomy perception and \textbf{(ii)} adaptive denoising.
%first
First, we develop a dose- and anatomy-aware contrastive language-image pre-training model~(\stageone) to achieve robust dose and anatomy perception by leveraging specialized contrastive learning strategies to learn continuous representations that quantify ordinal dose variations and identify salient anatomical regions.
%second
Second, we design a dose- and anatomy-aware diffusion model (\stagetwo) to perform adaptive and generalizable denoising by
synergistically integrating the learned dose and anatomy embeddings from \stageone into diffusion process via a novel dose and anatomy conditional block~(\block) based on Mamba.
%, which  through distinct mechanisms, each tailored to their respective characteristics.
% Central to \stagetwo is a novel dose and anatomy conditional block~(\block) that synergistically integrates dose embeddings via timestep-fused adaptive normalization and anatomical ones via a conditional state-space model (CSSM).
Extensive experiments on a large simulated multi-dose CT dataset spanning three anatomical regions, together with cross-dataset evaluations on Mayo-2016, CQ500, and piglet datasets, demonstrate superior denoising performance and strong generalization to unseen dose levels and anatomical regions.
The codes and models are available at \url{https://github.com/hao1635/FoundDiff}.
\end{abstract}


%% file: sections/introduction.tex
\section{Introduction}
\label{sec:introduction}
%background
\IEEEPARstart {C}{omputed} tomography (CT) is a widely-used imaging technique in clinical practice for diagnosis and screening. 
% , as it can provide accurate and detailed anatomical information~\cite{buzug2011computed}.
However, radiation exposure in normal-dose CT (NDCT) scans may cause unavoidable health damage and even increase the risk of cancers~\cite{smith2025projected}. 
% While low-dose CT (LDCT) reduces radiation exposure but often introduces noise and artifacts, affecting diagnostic accuracy. 
While low-dose CT (LDCT) reduces this risk, it introduces noise and artifacts that compromise diagnostic accuracy. 
% Consequently, efforts should be made to maintain radiation doses at levels that are as low as reasonably achievable (ALARA)\cite{shah2008alara} without compromising the diagnostic quality of the resulting images~\cite{dudhe2024radiation}.
Consequently, the goal is to minimize radiation exposure according to the as low as reasonably achievable (ALARA) principle~\cite{shah2008alara}, while preserving image quality for accurate diagnosis. 
% the diagnostic value of the images.

In recent years, deep learning (DL) techniques have been widely applied to suppress noise in LDCT imaging~\cite{chen2017low2,huang2021gan, Chi2025}. 
However, most existing DL-based methods are trained on data from a single dose level or a single anatomical region, which struggles to handle diverse noise characteristics and anatomical heterogeneity under varied scanning conditions~\cite{chen2023ascon}.
This constrains their generalizability and robustness in clinical practice, where CT scanning protocols vary due to different imaging requirements---such as patient age or target anatomy---and actual tube currents may differ across manufacturers even under identical nominal settings due to proprietary protocol implementations~\cite{need}.
A straightforward solution involves training specific models for such diverse dose and anatomy requirements.
However, this approach suffers from significant computation overhead and limited scalability.

\begin{figure}[t]
\centering
\includegraphics[width=1\linewidth]{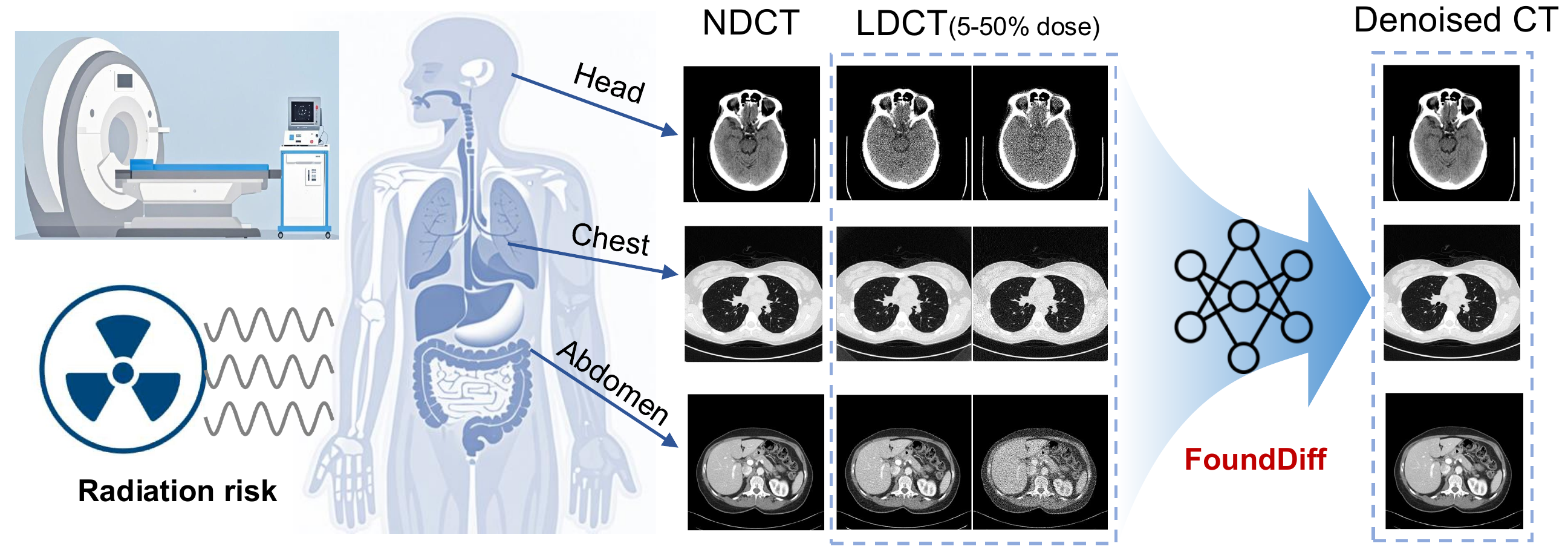}
\caption{Motivation for a unified LDCT denoising task.}
\label{fig:task}
\end{figure}

To address this challenge,  Liu~\etal pretrain a foundational imaging model on multiple CT image degradations, including diverse low-dose datasets~\cite{liu2024imaging}. 
However, their approach involves a straightforward integration of multiple tasks from large-scale datasets, which lacks adaptive image perception and requires fine-tuning when deployed on previously unseen data, thereby substantially increasing implementation and integration costs.
While efforts such as parameter-dependent framework (PDF)~\cite{xia2021pdf} have explored conditional training based on dose levels to enhance generalization, they rely on explicit and discrete input representations of dose, thus disregarding the continuous and ordinal nature of dose information and limiting their ability to effectively guide denoising for unseen data.
Furthermore, the methods aforementioned typically neglect inherent anatomical semantics during denoising, which is critical given that noise distribution varies according to anatomical regions~\cite{mussmann2021organ}.
Therefore, these challenges underscore the critical need for a foundational denoiser that perceives input characteristics and self-adaptively generalizes across both dose levels and anatomical variations without additional fine-tuning costs, shown in Fig.~\ref{fig:task}, which is crucial for enhancing the generalizability and practical clinical applicability of DL-based LDCT denoising technologies.
% While notable efforts, such as the parameter-dependent framework (PDF)~\cite{xia2021pdf}, have explored training across multiple dose levels with dose condition, they are constrained by the need for explicit dose information as input and 
% This limitation is critical because routine diagnostic exams cover varied regions whose noise statistics differ markedly owing to tissue-specific attenuation and density variations~\cite{chen2023ascon}.
% developing a unified denoising model that demonstrates robust adaptability across a wide spectrum of radiation doses and multiple anatomical regions remains a significant challenge. 
% Addressing this gap 

For this purpose, we propose \modelname, a foundational diffusion model that automatically perceives and adaptively denoises LDCT scans from diverse dose levels and anatomical regions; \blue{here, ``foundational''\infol{R1.2} in LDCT denoising means moving beyond “one model per setting” toward a unified framework.}
Our approach employs a two-stage learning strategy: \textbf{(i)} dose-anatomy perception and \textbf{(ii)} adaptive denoising. 
In the first stage, by the ability to encode a rich semantic feature space through contrastive language-image pre-training model~(CLIP)~\cite{clip}, we develop a dose- and anatomy-aware CLIP model, named \stageone, to rank ordinal dose levels and identify anatomy regions. 
Specifically, \stageone is built on an image quality assessment (IQA) framework of CLIPIQA~\cite{wang2023exploring} and fine-tuned on a large-scale CT dataset encompassing diverse dose levels and three key anatomical regions, \ie, abdomen, chest, and head, leveraging proposed dose ranking and anatomy discrimination losses. 
% to extract continuous ordinal feature of dose levels while concurrently identifying anatomical regions.
% Specifically, it involves two auxiliary loss components: a dose ranking contrastive loss to model ordinal nature of dose levels, and an anatomy-aware contrastive loss to discriminate between different anatomical regions.
By encoding the ordinal dose and clustered anatomy characteristics of diverse CT inputs through \stageone, the subsequent denoising process can be adaptively performed without explicit input condition, thereby enhancing the applicability in clinical settings.

%denoising
In the second stage, leveraging the advances of diffusion-based models in image restoration~\cite{liu2024rddm,gao2023corediff}, we propose a dose- and anatomy-aware diffusion model (\stagetwo), which performs adaptive and generalizable denoising by incorporating the learned representations from \stageone. 
In contrast to existing methods that apply a single, uniform fusion strategy for various types of priors~\cite{xia2021pdf}, the proposed \stagetwo leverages a novel dose and anatomy conditional block (\block) to employ distinct mechanisms tailored to the respective characteristics of dose and anatomical features for adaptive diffusion guidance: 
(\textbf{i})~dose embeddings are first fused with the timestep and then condition the module via adaptive normalization, suitable for modulating global noise statistics based on dose level; and 
(\textbf{ii})~anatomical embeddings are incorporated through a proposed conditional state-space model (CSSM), allowing focus on relevant anatomical information to enhance global contexts extraction with linear complexity. 
% This dual conditioning strategy enables robust adaptation across diverse dose levels and anatomical regions.
% Additionally, to facilitate interaction among the different channels of the CSSM, we incorporate transposed attention~\cite{zamir2022restormer,chen2024lit}, which performs self-attention along the channel dimension, also achieving linear complexity.

% Despite the immense popularity of the CNN~\cite{unet} or Transformer~\cite{transformer,vit,DiT} as the predominant architecture for diffusion models, it still exhibits substantial drawbacks of extensive memory and computational requirements.

% In contrast, recurrent-based state space models (SSM) in Mamba~\cite{gu2023mamba}, which summarize an arbitrarily long context in a single hidden state, can reduce these issues.
% Our \block integrates Mamba’s lightweight, efficient state-space models (SSMs) with Transformer’s powerful self-attention mechanism. 
% This hybrid design allows the model to effectively capture both local fine-grained features and long-range dependencies in the LDCT images.
% By doing so, \methodname achieve superior performace on publich dataset of eight differnt dose and three differnt antomical regions. Moreover, \methodname show remarkable generalization capybilites to unseen dose level, increasing the robustness on clincal scanieros. 
% 【第一次提continuous这个词，一直用这个吧，continous感觉才是单独一个感知模型的优势】

% and exhibits superior generalization capabilities across different datasets.
The main contributions of this work are listed as follows. 
First, we propose \modelname, a two-stage foundational diffusion model for unified and generalizable LDCT denoising across diverse dose levels and anatomical regions, without requiring additional conditions and costs, enhancing its clinical applicability.
Second, we propose \stageone, which can  achieve robust dose and anatomy perception by leveraging specialized contrastive learning strategies to predict ordinal dose levels and concurrently identify anatomical regions. 
Third, we propose \stagetwo, which can achieve adaptive denoising by fusing dose and anatomical information derived from \stageone into the diffusion process.
% and leverages transposed attention to enhance channel feature integration.
% \item Extensive experiments on public LDCT datasets and simulated data confirm that \modelname significantly outperforms competing methods across a range of dose levels and within three common anatomical sites. Crucially, the model displays remarkable generalization to previously unseen dose levels, highlighting its strong potential for widespread clinical adoption.
% Our model significantly improves the quality of denoised images, preserving crucial anatomical details and adapting seamlessly to multiple anatomical regions, making it highly suitable for broad clinical use.
Last, extensive experiments demonstrate that \modelname  outperforms state-of-the-art~(SOTA) methods for unified LDCT denoising, while maintaining robust generalization to unseen dose levels and anatomical regions.
%\end{enumerate}

% The remainder of this paper is organized as follows. 
% We first present the overall framework of the proposed \modelname in Sec.~\ref{sec:method}.
% Sec.~\ref{sec:experiments} provides comprehensive experimental results on diverse data, followed by  discussing the benefits and limitations of our method and some future works in Sec.~\ref{sec:discussion}.  Sec.~\ref{sec:conclusion} provides a concluding summary.

%% file: sections/method.tex
\begin{figure*}[h]
\centering
\includegraphics[width=1\linewidth]{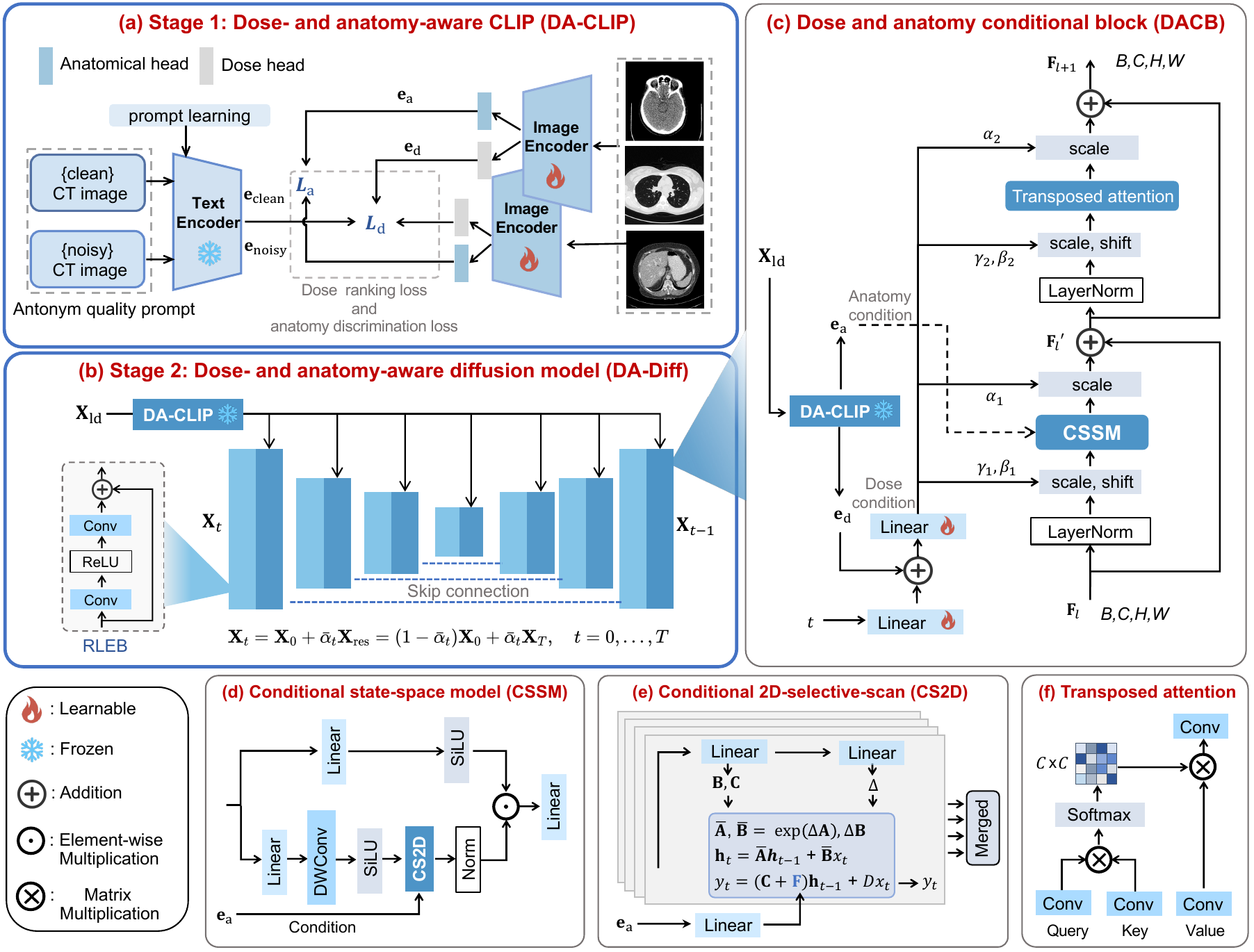}
\caption{Overview of \modelname. (a) Stage 1: dose- and anatomy-aware CLIP~(\stageone)  for dose-anatomy perception; (b) Stage 2: dose and anatomy-aware diffusion model~(\stagetwo) for adaptive denoising; (c) Dose and anatomy conditional block~(\block) within \stagetwo; (d-f) Key internal modules of the \block: (d) conditional state-space model (CSSM); (e) conditional 2D-selective-scan (CS2D); and (f) transposed attention.}
\label{networks}
\end{figure*}

\section{Method}
\label{sec:method}
Our study aims to develop a foundation model that unifies and generalizes LDCT denoising across diverse dose levels and anatomical regions, without requiring additional conditions and fine-tuning cost.
To this end, we propose a two-stage strategy comprising \stageone for dose-anatomy
perception and \stagetwo for adaptive denoising.
%Within \stagetwo, we design \block to effectively integrate dose and anatomy features from \stageone.
In the following, we first describe the overall framework and the hierarchical structure of \modelname in 
Subsec.~\ref{sec:overall_framework}. 
Then, we describe our two-stage model: \stageone in Subsec.~\ref{sec:DC-CLIP}, followed by \stagetwo and its core \block in Subsec.~\ref{sec:diffusion}.
% , followed by details of \stagetwo's core \block in Subsec.~\ref{sec:DMFormer}.

\subsection{Overall Framework of \modelname}
\label{sec:overall_framework}
Fig.~\ref{networks} presents the overview of the proposed \modelname framework consisting of two stages.
Fig.~\ref{networks}(a) presents the  first stage of dose and anatomy perception, where \stageone leverages the CLIPIQA framework~\cite{wang2023exploring} to capture both the dose-specific characteristics and anatomical features of CT images. 
Unlike traditional approaches that handle degradation and content features independently~\cite{gao2023great,AMIR}, \stageone unifies them within a single framework by fine-tuning a CLIP model with a dose ranking loss and an anatomical discrimination loss. 
This fine-tuning process enables the model to predict the ordinal dose level while identifying the anatomical region. 
% Specifically, prompt learning is introduced before the frozen text encoder to optimize the understanding of dose-related features, while the image encoder remains learnable. The rank loss ensures that the ordinal relationships between dose levels are captured, while the content preservation loss uses supervised contrastive learning to maintain the distinctiveness of anatomical regions.

Fig.~\ref{networks}(b) presents the second stage of adaptive denoising, where \stagetwo leverages a residual diffusion framework~\cite{liu2024rddm} to predict the residual image between NDCT and LDCT at each time step.
The \stagetwo follows a U-Net-like architecture with multiple encoder-decoder layers. 
Each level in the encoder and decoder consists of a residual local-enhance block (RLEB) and a \block. 
Input features first pass through the RLEB to capture local contextual information, then proceed to the core \block that is specifically designed to adaptively integrate dose and anatomical conditioning information extracted by \stageone to guide the denoising process.

%Next, we detail these two stages.
% The DMFormer block integrates two powerful components: the Selective State-space Model (SSM) from Mamba, which excels in modeling long-range dependencies with reduced computational complexity, and self-attention from the Transformer, which captures fine-grained local features. This hybrid design ensures that the model can efficiently handle the complex structures in LDCT images, making it well-suited for different anatomical regions such as the abdomen, chest, and head. Additionally, the integration of adaptive normalization layers and cross-attention mechanisms allows the model to adjust its behavior based on the dose and content embeddings, resulting in more effective noise reduction.

% In summary, the overall framework of UniCTD combines the strengths of DC-CLIP for dose and content representation with the dynamic denoising capabilities of the DMFormer block, enabling robust and adaptable LDCT denoising across diverse imaging scenarios.

\subsection{Stage 1: Dose- and Anatomy-aware CLIP (\stageone)}
\label{sec:DC-CLIP}

To effectively perceive dose and anatomical characteristics from diverse CT images, we introduce \stageone, as shown in Fig.~\ref{networks}(a).
Similar to describing image quality using text, \stageone is built upon a pre-trained CLIP-based IQA  framework~\cite{wang2023exploring}, which was adapted by fine-tuning the image encoder with two task-specific heads: a dose head and an anatomical head, designed to learn ordinal dose and discriminative anatomy embeddings, respectively.
To guide the fine-tuning process, we employ two specialized losses: a dose ranking loss and an anatomical discrimination loss, both of which are described in detail below.

\subsubsection{Dose ranking loss}
Accurate dose perception is critical for subsequent adaptive denoising~\cite{chen2023iqagpt}.
Previous CT IQA methods typically rely solely on images for classification or regression, supervised by a single label, and thus fail to capture the rich semantic structure of the feature space~\cite{gao2023great}. 
To this end, we incorporate CLIP by using natural language descriptions as supervision signals, enabling finer-grained semantic alignment and providing semantically enriched conditional information for subsequent adaptive denoising. 
This also advances the conventional regression models that learn discriminative representation rather than descriptive representation. 

% \noindent\textbf{Dose level prediction.}\quad
Specifically, the dose embedding, $\mat{e}_\text{d}$, is generated by the dose head of a two-layer MLP while $\mat{e}_\text{clean}$ and $\mat{e}_\text{noisy}$ are the text embeddings from the two prompts with opposite meanings, denoted by ``clean'' and ``noisy'' in our method.
The predicted dose level $\hat{y}_\mathrm{d}$ is calculated as follows:
% \begin{align}
% \label{eq:dose_score}
% \hat{y}_\mathrm{d}=\frac{\exp(\mat{e}_\text{d}\cdot \mat{e}_\text{clean})}{\exp(\mat{e}_\text{d}\cdot\mat{e}_\text{clean})+\exp(\mat{e}_\text{d}\cdot\mat{e}_\text{noisy})},
% \end{align}
% where $\boldsymbol{\cdot}$ symbol denotes the inner (dot) product. 
\begin{align}
\label{eq:dose_score}
\hat{y}_\mathrm{d}=\frac{\exp(\operatorname{sim_c}(\mat{e}_\text{d}, \mat{e}_\text{clean}))}{\exp(\operatorname{sim_c}(\mat{e}_\text{d}, \mat{e}_\text{clean}))+\exp(\operatorname{sim_c}(\mat{e}_\text{d}, \mat{e}_\text{noisy}))},
\end{align}
where $\operatorname{sim_c}$ denotes the cosine similarity.
Eq.~\eqref{eq:dose_score} indicates a softmax-normalized similarity between the dose embedding and the two opposite text embeddings. 
% When the pair of adjectives (``clean'' vs. ``noisy'') is used, the ambiguity of one prompt is reduced by its antonym as the task is cast as a binary classification.
Therefore, $\hat{y}_\mathrm{d}$ can be interpreted as estimating the posterior probability $p(\text{clean}\mid \text{CT}) \in [0,1]$, making it well-suited as a continuous quality score.
Moreover, from a latent-space perspective, $\mat{e}_\text{clean}$ and $\mat{e}_\text{noisy}$ act as two fixed anchors, so Eq.~\eqref{eq:dose_score} measures $\mat{e}_\text{d}$ by its relative similarity to these anchors, explicitly constraining the dose-related representation in the embedding space.
For dose level prediction, we employ MSE loss,  $\mathcal{L}_\text{dose}$, which is formulated as:
\begin{align}
\mathcal{L}_\text{dose}=\|\hat{y}_\mathrm{d}-y_\mathrm{d}\|_{2}^{2},  
\end{align}
where $y_\mathrm{d}$ is the label of dose level, defined as the relative fraction of the normal‐dose level used in the simulation (\eg, 1/4 and 1/10 of the normal dose).
Given that representations optimized with MSE loss often exhibit fragmentation and dispersion, relying solely on this objective is insufficient to capture the continuous and ordinal characteristics inherent to dose levels~\cite{zha2024rnc}.
This is especially problematic in LDCT imaging, where most dose labels cluster within the 10\%–50\% range and subtle differences are hard to distinguish.
% As the learned representations from MSE loss tend to exhibit fragmentation and fail to capture the continuous nature of sample ordering, leading to suboptimal performance in regression tasks~\cite{zha2024rnc}, 
To address this issue, we introduce a ranking contrastive loss $\mathcal{L}_\mathrm{rank}$, which reorganizes the images within a batch by their dose labels and contrasts them according to their relative orders. 
This ensures that the model learns the inherent ranking of the dose levels, improving performance on the ordinal dose level prediction.
The $\mathcal{L}_\mathrm{rank}$ is formulated as:
\begin{align}
%\small
\mathcal{L}_\mathrm{rank} = \frac{1}{2N({2N \!-\! 1})} \!\sum\limits_{i=1}^{2N} \!\sum\limits_{\substack{j=1 \\ j \neq i}}^{2N} \!-\! \log \frac{\exp\left(\frac{\mat{e}_\mathrm{d}^{i} \cdot\mat{e}_\mathrm{d}^{j}}{\tau}\right)}{\sum\limits_{\substack{\mat{e}_\mathrm{d}^{k}\in \mathcal{S}_{i,j}}} \exp\left(\frac{\mat{e}_\mathrm{d}^{i} \cdot\mat{e}_\mathrm{d}^{k}}{\tau}\right)},
\end{align}
where $S_{i,j} := \left\{ \mat{e}_\mathrm{d}^{k} \mid k \neq i, d(y_\mathrm{d}^i, y_\mathrm{d}^k) \geq d(y_\mathrm{d}^i, y_\mathrm{d}^j) \right\}$ represents the set of CT images that are ranked higher than $y_\mathrm{d}^j$ based on label distance relative to $y_\mathrm{d}^i$ and $d(\cdot,\cdot)$ is the $L1$ distance between two dose labels.
$\tau$ is the temperature parameter while $N$ is the batch size.
We follow the practice in~\cite{zha2024rnc} to set $\tau$ as 2.0 and treat the two augmented views of each sample as separate instances within a mini-batch, resulting in $2N$ views in total.
% Its objective is to learn a continuous representation by ranking these $2N$ instances based on the relative distances of their corresponding labels.}
This loss ensures that images with similar dose levels are mapped closer in the feature space, while those with significantly different dose levels are kept apart. 

Our final dose ranking loss involves the dose level prediction loss and ranking contrastive loss, which is defined as:
\begin{align}
\label{eq:L_dose}
\mathcal{L}_\mathrm{d} = \mathcal{L}_\mathrm{dose} + \mathcal{L}_\mathrm{rank}.
\end{align}

\subsubsection{Anatomy discrimination loss}
Previous studies typically focus on a single anatomical region during training, which causes some approaches~\cite{xia2021pdf} to condition only on dose levels while overlooking the anatomical semantic information. 
To simultaneously predict dose levels and capture anatomical features, we propose an anatomy discrimination loss based on supervised contrastive learning~\cite{supcon}, which ensures that features from different anatomical regions (\eg~abdomen, chest, and head) are well-clustered while distinguishing between various image features. 
The anatomy discrimination loss is defined as:
\begin{align}  \mathcal{L}_\mathrm{a} = \sum_{i=1}^{2N} \frac{-1}{|P(i)|} \sum_{p \in P(i)} \log \frac{\exp\left(\frac{\mat{e}_\mathrm{a}^{i} \cdot \mat{e}_\mathrm{a}^{p}}{ \tau}\right)}{\sum\limits_{j=1}^{2N} \exp\left(\frac{\mat{e}_\mathrm{a}^{i} \cdot\mat{e}_\mathrm{a}^{j}}{\tau}\right)},
\end{align}
where $\mat{e}_\mathrm{a}$ represents the anatomical embeddings computed by the anatomical head of a two-layer MLP, and $P(i)$ is the set of positive samples from the same anatomical region. 
$N$ is the batch size.
Consistent with Eq.~\eqref{eq:L_dose}, we adopt a two-view augmentation strategy, generating a total of $2N$ augmented views.
We set $\tau$ as 0.07,  a widely used default in contrastive learning, such as SimCLR~\cite{simclr} and the original SupCon~\cite{supcon}.
% This loss encourages the model to distinguish between different body regions and ensures that similar regions are grouped together in the feature space. 

% In the original CLIPIQA framework, both image and text encoders remain frozen, and the score in  Eq.~\eqref{eq:dose_score} serves as the output. 
% While extensive pre-training ensures proficiency on natural images, the baseline CLIPIQA struggles with the inherent complexities of medical images, particularly variations across dose levels and anatomy. 
% To enhance domain adaptation, CLIPIQA+\shan{no ref?} integrates prompt learning within the text encoder for improved image-text alignment, which is subsequently fine-tuned using the Mean Squared Error (MSE) loss.

\subsubsection{Objective function}
The overall loss function for \stageone, which integrates both ordinal dose level prediction and anatomical region identification, is defined as:
\begin{align}
\label{eq:L_daclip}
\mathcal{L}_\mathrm{\stageone}=\mathcal{L}_\mathrm{d}+\mathcal{L}_\mathrm{a}.
\end{align}

This loss guides the fine-tuning process, conducted on a large-scale CT dataset encompassing diverse dose levels and anatomical regions (detailed in Sec.~\ref{sec:experiments}).
Through this design, \stageone effectively captures both the dose- and anatomical-specific information for diverse CT images. This provides a robust foundation for building a unified CT denoising model, capable of more continuous and robust image perception without the explicit input condition,
thereby enhancing the applicability in diverse conditions. 
We empirically adopt equal weighting for simplicity in Eqs.~\eqref{eq:L_dose} and~\eqref{eq:L_daclip}. 
This choice is robust in practice and provides stable optimization, so we keep the default equal-weight design.

%DC-Diff
\subsection{Stage 2: Dose- and Anatomy-aware Diffusion Model~(\stagetwo)}
\label{sec:diffusion}
To effectively integrate the dose and anatomical information extracted from the \stageone and accurately guide the unified LDCT denoising, we present \stagetwo, shown in Fig.~\ref{networks}(b).
The denoising mechanism of \stagetwo is based on the RDDM framework~\cite{liu2024rddm}, which introduces a dual diffusion process that separates the denoising process into two components: noise diffusion and residual diffusion. 
The noise diffusion branch corresponds to conventional denoising diffusion probabilistic models (DDPMs)~\cite{DDPM} with stochastic Gaussian noise injection, while the residual diffusion branch belongs to the generalized diffusion family established by Cold Diffusion~\cite{bansal2023cold}, where deterministic forward degradation and learned reverse restoration serve as the core mechanism.
Following the setting of image restoration task in RDDM, we adopt the residual diffusion branch for LDCT denoising because it provides a directional and deterministic mapping between clean and degraded images~\cite{liu2024rddm}. 
In contrast, noise diffusion introduces additional stochasticity to encourage output diversity, which is undesirable for LDCT denoising that requires faithful and stable reconstructions. 
Prior studies also suggest that deterministic sampling typically achieves higher pixel-level fidelity than stochastic sampling~\cite{gao2023corediff,bansal2023cold}.

Given an LDCT $ \mat{X}_\mathrm{ld}$ and its corresponding NDCT $\mat{X}_\mathrm{nd}$. 
The residual image, denoted as:
$\mat{X}_\mathrm{res} = \mat{X}_\mathrm{ld} - \mat{X}_\mathrm{nd}$
is learned during the diffusion process, allowing the model to iteratively refine the image and reduce noise.
We denote the clean NDCT target as $\mat{X}_0 = \mat{X}_{\mathrm{nd}}$ and the degraded LDCT as $\mat{X}_T = \mat{X}_{\mathrm{ld}}$, such that $\mat{X}_{\mathrm{res}}=\mat{X}_T-\mat{X}_0$. The forward process is written as
\begin{align}
q(\mat{X}_t \mid \mat{X}_{t-1}, \mat{X}_\mathrm{res}) = \mathcal{N}(\mat{X}_t; \mat{X}_{t-1} + \alpha_t \mat{X}_\mathrm{res}, \beta_t^2 \mat{I}),
\end{align}
where $\alpha_t$ controls the step size along the residual direction and $\beta_t$ controls the injected noise. 
$\mathbf{I}$ denotes the identity matrix.
Since we adopt residual diffusion only, we set $\beta_t=0$ for all $t$, and the forward process degenerates to a deterministic interpolation. 
Defining $\bar{\alpha}_t:=\sum_{s=1}^{t}\alpha_s$ with $\bar{\alpha}_0=0$ and $\bar{\alpha}_T=1$, we obtain the closed-form expression:
% \begin{align}
% \mat{X}_t \mathrel{\scriptstyle=} \mat{X}_0 \mathrel{\scriptstyle+} \bar{\alpha}_t \mat{X}_{\mathrm{res}}
% \mathrel{\scriptstyle=} (1-\bar{\alpha}_t) \mat{X}_0 \mathrel{\scriptstyle+} \bar{\alpha}_t \mat{X}_T,
% t\mathrel{\scriptstyle=}0,\ldots,T.
% \label{eq}
% \end{align}
\begin{align}
\mat{X}_t \mathrel{\scalebox{0.65}{$=$}} \mat{X}_0 + \bar{\alpha}_t \mat{X}_{\mathrm{res}}
\mathrel{\scalebox{0.65}{$=$}} (1-\bar{\alpha}_t) \mat{X}_0 + \bar{\alpha}_t \mat{X}_T,
t\mathrel{\scalebox{0.65}{$=$}}0,\ldots,T.
\label{eq}
\end{align}
Given the deterministic forward formulation, the reverse process starts from $\mat{X}_T$ and progressively recovers the clean NDCT image by subtracting the predicted residual increment at each step:
\begin{align}
\mat{X}_{t-1} = \mat{X}_t - \alpha_t\mathcal{D}_{\theta}(\mat{X}_t, t, \mat{X}_\mathrm{ld}),
\quad t=T,\ldots,1,
\label{eq}
\end{align}
where $\mathcal{D}_{\theta}(\cdot)$ is the DA-Diff denoising network conditioned on the diffusion timestep $t$. 
The final output $\hat{\mat{X}}_0$ is obtained after $T$ reverse steps.

The denoising network primarily comprises RLEB and dose and anatomy conditional blocks~(\block) shown in Fig.~\ref{networks}(b). 
Input features at each level first pass through the RLEB before entering \block. 
The \block integrates timestep and \stageone derived dose and anatomy conditional information to guide the diffusion process, which is detailed below.

\subsubsection{Dose and anatomy conditional block~(\block)}
\label{sec:DMFormer}
\block is the core design in \stagetwo, enabling the effective integration of dose and anatomical embeddings from \stageone, as shown in Fig.~\ref{networks}(c). 
It takes as input the intermediate feature map $\mat{F}_l$, the dose embedding $\mat{e}_\mathrm{d}$, the anatomical embedding $\mat{e}_\mathrm{a}$, and the diffusion timestep $t$ as input, and outputs a refined feature map $\mat{F}_{l+1}$.
Unlike prior approaches that use a single fusion strategy for all priors~\cite{xia2021pdf}, \stagetwo employs the proposed \block to apply distinct mechanisms tailored to the characteristics of dose and anatomy for adaptive diffusion guidance.

\noindent\textbf{Dose conditioning via timestep fusion.}\quad  To capture the joint influence of the image dose level and the current step of the diffusion process, we fuse the dose embedding $\mat{e}_\mathrm{d}$ and the time step $t$ through an adaptive layer normalization with zero initialization~(adaLN-Zero) operation~\cite{DiT}. 
The motivation behind this design is that both dose level and diffusion timestep represent global image states, making adaptive normalization suitable for modulating feature statistics globally based on these combined signals.
Specifically, the dose and timestep embeddings are first fused via summation after processing $t$ with a linear layer and then fed into a linear layer to predict six zero-initialized modulation parameters: $\gamma_1, \beta_1, \alpha_1,\gamma_2, \beta_2, \alpha_2$, where $\alpha$, $\gamma$ are the scaling parameters and $\beta$ is the shifting parameter.
These parameters are dynamically generated to adjust the behavior of the normalization layers, which are produced by:
\begin{align}
\gamma_1, \beta_1, \alpha_1,\gamma_2, \beta_2, \alpha_2=\operatorname{MLP}(\operatorname{MLP}(t)+\mat{e}_\mathrm{d}).
\end{align}
This allows the normalization layer to adapt to the input dose characteristics and the progression of the diffusion process, ensuring effective and adaptive LDCT denoising.

\noindent\textbf{Anatomy conditioning via conditional state-space
model (CSSM).}\quad To incorporate the specific anatomical spatial information, we propose CSSM, whose architecture is depicted in Fig.~\ref{networks}(d). 
CSSM is based on the state space model (SSM) in Mamba architecture~\cite{gu2023mamba}, known for capturing global context efficiently with linear complexity.
As shown in Fig.~\ref{networks}(e), SSM is rooted in linear time-invariant~(LTI) systems~\cite{gu2021efficiently}, which transform a sequence $x(t) \in \mathbb{R}$ to $y(t) \in \mathbb{R}$ via an implicit latent state $\mathbf{h}(t) \in \mathbb{R}^{M}$, governed by a linear ordinary differential equation (ODE):
\begin{equation}
\begin{aligned}
\label{eq:ssm}
    \mat{h^{\prime}}(t)&={\mat{A}}\mat{h}(t)+{\mat{B}}{x(t)},\\
    y(t)&={\mat{C}}\mat{h}(t)+{D}{x(t)},
\end{aligned}
\end{equation}
where $M$ is the state size, ${\mat{A}} \in \mathbb{R}^{M\times M}$, ${\mat{B}} \in \mathbb{R}^{M \times 1}$, ${\mat{C}} \in \mathbb{R}^{1\times M}$, and ${D} \in \mathbb{R}$. 
After discretization, the continuous SSM can be rewritten as a convolution for faster training.
S4~\cite{gu2021efficiently} improves efficiency via structured SSM parameter matrices.
Mamba further introduces selective scanning for input-dependent updates and efficient sequence computation.
For vision tasks, 2D selective scan~(SS2D)~\cite{liu2024vmamba, pei2024efficientvmamba} extends selective scanning from 1D sequences to 2D images by scanning spatial directions, capturing long-range dependencies.

To incorporate anatomical conditions, we introduce CS2D, a conditional variant of the SS2D, shown in Fig.~\ref{networks}(e), where the anatomical embedding $\mat{e}_\mathrm{a}$ undergoes a linear projection before being added to the SSM's output matrix $\mat{C}$. 
By incorporating the conditional scanning mechanism in CSSM, the \block effectively enhances long-range dependencies and integrates anatomically relevant spatial information.
Recognizing that standard SSMs process channels independently, potentially leading to redundancy~\cite{guo2024mambair}, we add transposed attention~\cite{chen2024lit} into the \block to facilitate channel interaction. 
It computes self-attention across the channel dimension from the transposed query and key to obtain the attention map ($C\times C$, $C$ is the number of channels), enabling the learning of diverse channel representations and mitigating redundancy.

\noindent\textbf{Forward pass in DACB.}\quad 
In general, the \block is structured as follows: 
Given the feature $\mathbf{F}_{l}$ at layer $l$, the input is first passed through a LayerNorm operation~\cite{lei2016layernorm}, followed by scaling with the parameters $\gamma_1$ and $ \beta_1$.
This is then processed by the CSSM with anatomical condition $\mat{e}_\mathrm{a}$, which integrates anatomical information while efficiently capturing long-range dependencies. 
The output from the CSSM is subsequently combined with a residual connection, scaled by $\alpha_1$. This process can be formalized as:
\begin{align}
% \mat{F}_{\text{norm1}} &= \text{LayerNorm}(\mat{F}_l) \\
\mat{F}_{\text{mod1}} &= (1 + \gamma_1) \cdot \text{LayerNorm}(\mat{F}_l) + \beta_1 \\
\mat{F}_l^{\prime} &= \mat{F}_l + \alpha_1 \cdot \text{CSSM}(\mat{F}_{\text{mod1}}, \mat{e}_\mathrm{a}).
\label{eq:R2_9_eq1}
\end{align}
After the CSSM operation, the second instance of adaLN-Zero is applied with the transposed attention mechanism. 
Here, we further modulate the feature using the second set of parameters:
\begin{align}
% \mat{F}_{\text{norm2}} &=  \\
\mat{F}_{\text{mod2}} &= (1 + \gamma_2) \cdot \text{LayerNorm}(\mat{F}_l^{\prime}) + \beta_2 \\
\mat{F}_{l+1} &= \mat{F}_l^{\prime}  + \alpha_2 \cdot \text{TransposedAttention}(\mat{F}_{\text{mod2}}).
\label{eq:R2_9_eq2}
\end{align}
% given the feature $\mathbf{F}_{l}$ at layer $l$, the input is first passed through a LayerNorm operation~\cite{lei2016layernorm}, followed by scaling with the parameters $\gamma_1$ and $ \beta_1$.
% This is then processed by the CSSM with anatomical condition $\mat{e}_\mathrm{a}$, which integrates anatomical information while efficiently capturing long-range dependencies. 
% The output from the CSSM is subsequently combined with a residual connection, scaled by $\alpha_1$. This process can be formalized as:
% \begin{align}
% \mathbf{F}_l^1&=\gamma_1  \mathrm{LayerNorm}(\mathbf{F}_l) + \beta_1
% \mathbf{F}_l \quad\quad \\
% \mathbf{F}_{l}^{\prime} &= \mathrm{CSSM}(\mathbf{F}_l^1, \mat{e}_\mathrm{a}) + \alpha_1 \mathbf{F}_l^1+\mathbf{F}_l.
% \end{align}
% After the CSSM operation, the second instance of adaLN-Zero is applied with the transposed attention mechanism. 
% Here, we further modulate the feature using the second set of parameters 
% $\alpha_2$, $\gamma_2$, and $\beta_2$, computed as:
% \begin{align}
% \mathbf{F}_l^2 &=\gamma_2  \mathrm{LayerNorm}(\mathbf{F}_l^{\prime}) + \beta_2
% \mathbf{F}_l^{\prime} \quad\quad\quad\quad \\
% \mathbf{F}_{l+1} &= \operatorname{TransposedAttention}(\mathbf{F}_{l}^{\prime})+\alpha_2 \mathbf{F}_l^2+\mathbf{F}_l^{\prime}.
% \end{align}
The final output feature $\mathbf{F}_{l+1}$ incorporates both dose information from the adaLN-Zero and anatomical region contexts from the CSSM mechanism, modulated by the dose and anatomical embeddings from \stageone, respectively. 
This allows the \stagetwo to adapt its denoising process based on both dose level and anatomical information.

\subsubsection{Objective function}

\stagetwo is trained using a residual loss function, which focuses on predicting the $\mat{X}_\mathrm{res}$. This ensures the model learns to progressively refine the input image at each timestep, producing cleaner results. The training loss of \stagetwo is defined as:
\begin{align}
\mathcal{L}_\mathrm{\stagetwo}(\theta) = \mathbb{E} \left[\lVert \mat{X}_\mathrm{res} - \mat{X}_\mathrm{res}^\theta(\mat{X}_t, t, \mat{X}_\mathrm{ld}) \rVert^2 \right].
\end{align}
The expectation $\mathbb{E}$ is computed over the distribution of training data and noise levels. 
Minimizing the $\mathcal{L}_\mathrm{\stagetwo}$ within the residual diffusion mechanism allows \stagetwo to achieve high-quality reconstruction with reduced inference time.

%\subsubsection{Timestep-fused dose conditioning}

%\subsubsection{Anatomical conditioning via CSSM}

% Furthermore, transposed attention can supplement the global context extraction, doing so with linear complexity.

%% file: sections/experiment.tex
\section{Experiments}
\label{sec:experiments}

\subsection{Datasets}
\label{sec:dataset}
\subsubsection{Large simulated dataset}
To effectively train the proposed \modelname, we simulate a large-scale dataset encompassing a variety of dose levels and anatomical regions. 
This simulation is based on the widely recognized LDCT images and projection dataset~\cite{moen2021low}, referred to as the Mayo-2020 dataset. 
The dataset consists of CT scans and their corresponding LDCT images from 50 patients for each anatomical region---abdomen, chest, and head---acquired using GE Discovery and Siemens SOMATOM scanners. 
In the original dataset, chest LDCT images are acquired at 1/10 of the normal dose, while abdomen and head LDCT images are acquired at 1/4 of the normal dose.
To expand the range of dose levels suitable for our task, we employ the ASTRA Toolbox~\cite{van2016fast,van2015astra} to simulate ten distinct dose levels: 1/2, 1/3, 1/4, 1/5, 1/6, 1/7, 1/8, 1/10, 1/15 and 1/20.
The imaging system  uses 1,152 views and 672 detector bins with a detector bin size of 1.85 mm. The distance from the X-ray source to the rotation center is 746.02 mm, and the distance from the detector to the rotation center is 615.18 mm. Images are reconstructed on a $512\times512$ grid. 
% $1/2$, $1/3$, $1/4$, $1/5$, $1/6$, $1/8$, $1/10$, and $1/20$. 
% The ASTRA Toolbox is well-established for its accuracy in low-dose simulation, ensuring that the generated data closely mimics the noise characteristics observed in real-world LDCT scans.
To generate the noise level of the original datasets as closely as possible, we adopt a widely used \emph{Poisson\,+\,Gaussian} measurement model in the intensity domain and calibrate the baseline incident photon count per bin per view, $I_0$, by anatomy.
The Poisson sampling is implemented using ASTRA's provided Poisson-noise function.
Given a normal-dose projection $p_{\mathrm{hd}}$, the low-dose projection for anatomy $a\in\{\text{abdomen}, \text{chest}, \text{head}\}$ is
\begin{equation}
\label{eq:poissongauss}
p_{\mathrm{ld}}^{(a)}
= \ln\!\left(
\frac{I^{(a)}}{\mathrm{Poisson}\!\big(I^{(a)} \exp(-p_{\mathrm{hd}})\big) + \mathcal{N}(0,\sigma_e^2)}
\right),
\end{equation}
where $I^{(a)} = y_\mathrm{d} I_0^{(a)}$ and $y_\mathrm{d}$ denotes the dose fraction (with the ten dose levels specified above)
and a fixed electronic-noise variance $\sigma_e^2=10$~\cite{xia2021pdf}. We empirically calibrate the incident photon count $I_0$ to match the noise and texture of Mayo-2020 LDCT.
Specifically, we set $I_0^{(\text{abdomen})} = I_0^{(\text{chest})} = 1.5 \times 10^{5}$ and $I_0^{(\text{head})} = 4.5 \times 10^{6}$. 
We do not model system blurring; instead, we focus on dose-dependent quantum and electronic noise as in Eq.~\eqref{eq:poissongauss}.
For reconstruction, all sinograms are reconstructed using the filtered back projection (FBP) algorithm.
Prior studies~\cite{zeng2015simple,gao2023corediff} have shown that LDCT can be simulated by scaling the incident photon flux with a dose fraction $y_\mathrm{d}$ and modeling the resulting measurements using Poisson noise plus additive electronic noise, which closely matches real low-dose acquisitions.
We retain the originally provided LDCT images in the Mayo-2020 dataset and simulate only the remaining dose levels to complete the dataset.
Finally, we obtain 13,363 CT slices per dose level, with 5,339/6,616/1,408 slices from abdomen/chest/head, respectively. Since we simulate ten dose levels for each slice, the overall number of slices across all dose levels is 133,630.
We select 80\% of the patients for training and the rest for testing.

\subsubsection{Mayo-2016 dataset}
To evaluate the generalization performance of different methods to  unseen data, we use the 2016 AAPM Grand Challenge dataset~\cite{mccollough2017low}, referred to as the Mayo-2016 dataset, which includes abdominal CT images from 10 patients. Each scan is acquired using a Siemens SOMATOM Flash scanner and reconstructed with a B30 kernel. For dose-level testing, we randomly selected two patients, yielding a total of 1,136 images, in a manner similar to previous studies~\cite{chen2024lit, chen2023ascon}. 
Although collected at the same center, the Mayo-2016 and Mayo-2020 datasets differ in key aspects of the LDCT simulation workflow~\cite{mccollough2017low,moen2021low}. Therefore, Mayo-2016 serves as a complementary dataset to explore robustness to a shifted noise distribution from the same center.

\subsubsection{CQ500 dataset} 
To evaluate generalization across different scanners and institutions, we utilized the public CQ500 dataset~\cite{cq500}. 
CQ500 is a multi-center head CT dataset from six radiology centers in New Delhi, India, acquired on multiple GE and Philips scanners~\cite{cq500}.
We randomly select 20 patients with 658 slices. 
Since this dataset only provides NDCT, we simulate four dose levels: 1/4, 1/7, 1/10 and 1/15. 
The LDCT simulation procedure is identical to our large simulated dataset.

\subsubsection{Piglet dataset} 
To further examine cross-anatomical generalization on real low-dose scans, we additionally test \modelname on the public piglet dataset~\cite{piglet}. 
CT scans were acquired on a GE Discovery CT750 HD scanner with reduced tube current, including one normal-dose scans (300~mAs) and four low-dose scans at 1/2, 1/4, 1/10, and 1/20 dose levels (150/75/30/15~mAs)~\cite{piglet}.
Each dose level contains 850 paired NDCT/LDCT images with real acquisition noise~\cite{piglet}.
Importantly, piglet anatomy and tissue characteristics are unseen in human training data, providing a stringent cross-anatomy test.

\subsection{Implementation Details}\label{sec:implementation}
For the training of \stageone, the network is trained for 100 epochs with a batch size of 192 on an NVIDIA V100 GPU. We utilize the stochastic gradient descent (SGD) optimizer \cite{loshchilov2017decoupled}, setting the momentum to 0.9 and weight decay to $1.0\times10^{-9}$. The learning rate is initialized at $1.0\times 10^{-2}$ and reduced progressively to $1.0\times 10^{-5}$ using a cosine annealing schedule~\cite{loshchilov2016sgdr}.

For the \stagetwo, training is performed for 400K iterations with a batch size of 2 on one NVIDIA RTX 3090 GPU (24GB). 
The Adam optimizer \cite{loshchilov2017decoupled} is used, with momentum parameters set to $\beta_{1}=0.9$, $\beta_{2}=0.99$, and a weight decay of $1.0\times10^{-9}$. The learning rate is initialized at $2.0\times 10^{-4}$.
The denoising network in \stagetwo consists of four levels with channel dimensions of 64, 128, 256, and 512, respectively. 
The total number of diffusion steps ($T$) is set to $1\times 10^{3}$. 
We empirically set the total sampling steps to 2, which provides the best trade-off between denoising quality and efficiency; this setting is also consistent with the baseline RDDM configuration, ensuring a fair comparison.

Regarding data preprocessing, for \stageone, image patches of size 256$\times$256 are used, while the \stagetwo operates on full-sized images of 512$\times$512, with a window level set to [-1000, 2000]~HU for all models.
Data augmentation is performed by randomly applying horizontal flips and rotating the images by $90^{\circ}$, $180^{\circ}$, and $270^{\circ}$.

For quantitative evaluation, \stageone is assessed using the Pearson linear correlation coefficient (PLCC) and Spearman's rank-order correlation coefficient (SROCC). The denoising performance is evaluated using two standard metrics: peak signal-to-noise ratio (PSNR) and structural similarity index measure (SSIM)~\cite{wang2004image}.
As a complement, we additionally report four perceptual and structure-aware metrics on the large simulated dataset, including learned perceptual image patch similarity (LPIPS)~\cite{zhang2018unreasonable}, visual information fidelity (VIF)~\cite{sheikh2006image}, feature similarity index~(FSIM)~\cite{zhang2011fsim}, and gradient magnitude similarity deviation (GMSD)~\cite{xue2013gradient}.

\subsection{Competing Methods}
To evaluate the performance of the proposed \modelname, three types of methods are selected for comparison, including 1) Baseline method: RDDM~\cite{liu2024rddm}; 2) SOTA methods for LDCT denoising: RED-CNN~\cite{chen2017low2}, PDF-based RED-CNN~\cite{xia2021pdf}, and CoreDiff~\cite{gao2023corediff}; 
3) SOTA methods for universal image restoration: Restormer~\cite{zamir2022restormer} for natural images and AMIR~\cite{AMIR} for medical images. 
PDF-based RED-CNN is a conditional network based on RED-CNN, where the dose levels are parameterized and fed into two MLPs; we name it PDF in this section. 
CoreDiff and RDDM are both diffusion-based methods.
For the RDDM baseline, we configure the model to operate only in residual-diffusion mode to ensure a strictly fair comparison.
AMIR is a universal medical image restoration method using a task-adaptive routing strategy.

While prior-based unsupervised approaches~\cite{hein2024physics,liu2025diffusion} can generalize across imaging conditions via diffusion priors, their performance still trails SOTA paired supervised approaches~\cite{need}. 
Therefore, since our goal is to benchmark paired supervised methods for unified LDCT imaging, we exclude unsupervised baselines from the main comparison.
% We nonetheless discuss these methods to in Sec.~\ref{sec:discussion}.}

To ensure a fair and reproducible comparison, we use the official public implementations of all competing methods and keep their original paper-reported hyperparameters;
moreover, all methods are trained and evaluated using the same GPU configuration as \modelname.
We also match the overall training budget by aligning the total data usage (batch size × iterations) of each competitor to that of our \modelname, so that no method is under-trained. 
For evaluation, we report results from the final converged checkpoint rather than selecting the best epoch on the validation set, which avoids test-set leakage.

% \begin{figure}[!h]
% \centering
% \includegraphics[width=1\linewidth]{figures/seendata_radar_final.pdf}
% \caption{Quantitative results of PSNR and SSIM on four seen dose levels from the large simulated dataset.}
% \label{fig:simdata_seen_radar}
% \end{figure}

% \begin{figure*}[h]
% \centering
% \includegraphics[width=1\linewidth]{figures/alldata_radar_final.pdf}
% \caption{Quantitative results of PSNR and SSIM on four unseen dose levels from the large simulated dataset.}
% \label{fig:simdata_unseen_radar}
% \end{figure*}

\begin{table*}[t]
\centering
% \scriptsize
% \setlength{\tabcolsep}{1.5pt}
%\renewcommand\arraystretch{1}
\caption{Quantitative comparison on \textbf{SEEN} dose levels in terms of PSNR and SSIM [$\times 10^{-2}$]. Best results are in \textbf{bold}.}
\label{tab:seen_results}
\resizebox{\textwidth}{!}{
\begin{tabular}{c|l|cc|cc|cc|cc}
\shline
\multirow{2}{*}{\textbf{Anatomy}} & \multirow{2}{*}{\textbf{Method}} &
\multicolumn{2}{c|}{\textbf{1/2 Dose}} &
\multicolumn{2}{c|}{\textbf{1/4 Dose}} &
\multicolumn{2}{c|}{\textbf{1/6 Dose}} &
\multicolumn{2}{c}{\textbf{1/10 Dose}} \\
\cline{3-4} \cline{5-6} \cline{7-8} \cline{9-10}
& & PSNR$\uparrow$ & SSIM$\uparrow$ & PSNR$\uparrow$ & SSIM$\uparrow$ & PSNR$\uparrow$ & SSIM$\uparrow$ & PSNR$\uparrow$ & SSIM$\uparrow$ \\
\hline

% --- Abdomen Region ---
\multirow{7}{*}{\textbf{Abdomen}}
& RED-CNN   & 47.37{$\pm$0.95} & 98.53{$\pm$0.39} & 47.96{$\pm$2.34} & 99.19{$\pm$0.16} & 44.94{$\pm$0.54} & 97.59{$\pm$0.23} & 44.24{$\pm$0.90} & 97.46{$\pm$0.31} \\
& PDF       & 47.35{$\pm$0.77} & 98.46{$\pm$0.30} & 47.90{$\pm$2.22} & 99.21{$\pm$0.16} & 45.17{$\pm$0.79} & 97.57{$\pm$0.48} & 44.18{$\pm$1.24} & 97.33{$\pm$0.56} \\
& Restormer & 48.37{$\pm$0.73} & 98.99{$\pm$0.17} & 48.15{$\pm$2.24} & 99.23{$\pm$0.16} & 46.37{$\pm$0.79} & 98.55{$\pm$0.21} & 45.13{$\pm$1.05} & 98.16{$\pm$0.31} \\
& AMIR      & 48.52{$\pm$0.72} & 98.98{$\pm$0.17} & 48.08{$\pm$2.42} & 99.05{$\pm$0.24} & 46.54{$\pm$0.81} & 98.55{$\pm$0.22} & 45.29{$\pm$1.09} & 98.18{$\pm$0.35} \\
& CoreDiff  & 48.51{$\pm$0.77} & 98.98{$\pm$0.17} & 48.33{$\pm$2.07} & 99.23{$\pm$0.16} & 46.56{$\pm$0.79} & 98.54{$\pm$0.19} & 45.34{$\pm$1.00} & 97.18{$\pm$1.25} \\
& RDDM      & 48.70{$\pm$0.78}    & 99.05{$\pm$0.16}    & 48.65{$\pm$2.16}    & 99.25{$\pm$0.18}    & 46.69{$\pm$0.84}    & 98.63{$\pm$0.22}    & 45.43{$\pm$1.11}    & 98.21{$\pm$0.36} \\
\rowcolor{cyan!20}\cellcolor{white}&  \textbf{FoundDiff} & \textbf{48.92{$\pm$0.74}} & \textbf{99.12{$\pm$0.15}} & \textbf{48.88{$\pm$2.34}} & \textbf{99.32{$\pm$0.17}} & \textbf{47.02{$\pm$0.79}} & \textbf{98.77{$\pm$0.19}} & \textbf{45.91{$\pm$0.95}} & \textbf{98.49{$\pm$0.24}} \\
\hline

% --- Chest Region ---
\multirow{7}{*}{\textbf{Chest}}
& RED-CNN   & 36.44{$\pm$1.96} & 86.56{$\pm$4.68} & 35.86{$\pm$1.87} & 84.85{$\pm$4.93} & 35.54{$\pm$1.83} & 83.72{$\pm$5.14} & 34.20{$\pm$2.69} & 81.82{$\pm$6.55} \\
& PDF       & 36.24{$\pm$1.88} & 85.69{$\pm$4.31} & 35.80{$\pm$1.83} & 84.54{$\pm$4.70} & 35.43{$\pm$1.82} & 82.85{$\pm$5.00} & 34.01{$\pm$2.79} & 81.49{$\pm$6.54} \\
& Restormer & 36.55{$\pm$1.99} & 86.85{$\pm$5.03} & 35.96{$\pm$1.91} & 85.06{$\pm$5.31} & 35.66{$\pm$1.88} & 84.11{$\pm$5.52} & 34.24{$\pm$2.71} & 82.07{$\pm$6.37} \\
& AMIR      & 36.60{$\pm$1.94} & 86.84{$\pm$4.63} & 35.99{$\pm$1.88} & 85.04{$\pm$4.98} & 35.68{$\pm$1.86} & 84.07{$\pm$5.27} & 34.32{$\pm$2.71} & 81.79{$\pm$6.92} \\
& CoreDiff  & 36.48{$\pm$1.92} & 87.02{$\pm$4.85} & 35.90{$\pm$1.92} & 84.93{$\pm$5.41} & 35.61{$\pm$1.89} & 83.70{$\pm$5.56} & 34.16{$\pm$2.74} & 81.74{$\pm$6.21} \\
& RDDM      & 36.43{$\pm$1.85}    & 87.06{$\pm$4.07}    & 35.78{$\pm$1.85}    & 84.99{$\pm$4.90}    & 35.45{$\pm$1.83}    & 83.92{$\pm$5.26}    & 33.81{$\pm$2.71}    & 81.01{$\pm$6.55} \\
\rowcolor{cyan!20}\cellcolor{white}&  \textbf{FoundDiff} & \textbf{36.82{$\pm$1.82}} & \textbf{88.10{$\pm$3.65}} & \textbf{36.14{$\pm$1.85}} & \textbf{85.92{$\pm$4.63}} & \textbf{35.81{$\pm$1.85}} & \textbf{84.78{$\pm$5.16}} & \textbf{34.36{$\pm$2.79}} & \textbf{82.13{$\pm$6.35}} \\
\hline

% --- Head Region ---
\multirow{7}{*}{\textbf{Head}}
& RED-CNN   & 51.77{$\pm$6.53} & 99.34{$\pm$0.58} & 55.13{$\pm$4.30} & 99.65{$\pm$0.87} & 50.85{$\pm$6.14} & 99.25{$\pm$0.60} & 50.28{$\pm$5.87} & 99.21{$\pm$0.59} \\
& PDF       & 51.02{$\pm$6.32} & 99.25{$\pm$0.52} & 55.10{$\pm$4.36} & 99.55{$\pm$1.26} & 50.09{$\pm$5.93} & 99.16{$\pm$0.53} & 49.17{$\pm$5.68} & 99.01{$\pm$0.62} \\
& Restormer & 53.59{$\pm$6.52} & 99.61{$\pm$0.42} & 55.74{$\pm$4.28} & 99.72{$\pm$0.48} & 52.33{$\pm$6.07} & 99.54{$\pm$0.45} & 51.57{$\pm$5.84} & 99.48{$\pm$0.46} \\
& AMIR      & 53.14{$\pm$6.29} & 99.28{$\pm$0.54} & 55.41{$\pm$4.09} & 99.67{$\pm$0.53} & 52.35{$\pm$6.05} & 99.38{$\pm$0.52} & 51.65{$\pm$5.85} & 99.36{$\pm$0.51} \\
& CoreDiff  & 53.56{$\pm$6.51} & 99.69{$\pm$0.46} & 54.34{$\pm$4.28} & 99.60{$\pm$0.75} & 52.25{$\pm$6.07} & 99.51{$\pm$0.48} & 51.46{$\pm$5.83} & 99.43{$\pm$0.50} \\
& RDDM      & 55.02{$\pm$6.02}    & 99.69{$\pm$0.38}    & 56.15{$\pm$4.31}    & 99.73{$\pm$0.54}    & 53.50{$\pm$5.54}    & 99.62{$\pm$0.40}    & 52.63{$\pm$5.33}    & 99.56{$\pm$0.42} \\
\rowcolor{cyan!20}\cellcolor{white}& \textbf{FoundDiff} & \textbf{55.59{$\pm$6.18}} & \textbf{99.74{$\pm$0.36}} & \textbf{56.67{$\pm$4.38}} & \textbf{99.76{$\pm$0.48}} & \textbf{53.92{$\pm$5.67}} & \textbf{99.67{$\pm$0.39}} & \textbf{53.03{$\pm$5.45}} & \textbf{99.62{$\pm$0.42}} \\
\shline
\end{tabular}}
\end{table*}

\begin{table*}[t]
\centering
% \scriptsize
% \setlength{\tabcolsep}{1.5pt}
\renewcommand\arraystretch{1}
\caption{Quantitative comparison on \textbf{UNSEEN} dose levels in terms of PSNR and SSIM[$\times 10^{-2}$]. Best results are in \textbf{bold}.}
\label{tab:unseen_results}
\resizebox{\textwidth}{!}{
\begin{tabular}{c|l|cc|cc|cc|cc}
\shline
\multirow{2}{*}{\textbf{Anatomy}} & \multirow{2}{*}{\textbf{Method}} &
\multicolumn{2}{c|}{\textbf{1/5 Dose}} &
\multicolumn{2}{c|}{\textbf{1/7 Dose}} &
\multicolumn{2}{c|}{\textbf{1/15 Dose}} &
\multicolumn{2}{c}{\textbf{1/20 Dose}} \\
\cline{3-4} \cline{5-6} \cline{7-8} \cline{9-10}
& & PSNR$\uparrow$ & SSIM$\uparrow$ & PSNR$\uparrow$ & SSIM$\uparrow$ & PSNR$\uparrow$ & SSIM$\uparrow$ & PSNR$\uparrow$ & SSIM$\uparrow$ \\
\hline

% --- Abdomen Region ---
\multirow{7}{*}{\textbf{Abdomen}}
& RED-CNN   & 45.14{$\pm$0.58} & 97.63{$\pm$0.25} & 44.79{$\pm$0.56} & 97.59{$\pm$0.24} & 42.78{$\pm$2.35} & 96.67{$\pm$1.13} & 41.06{$\pm$3.76} & 95.11{$\pm$2.64} \\
& PDF       & 45.14{$\pm$0.82} & 97.17{$\pm$0.69} & 44.86{$\pm$0.83} & 97.47{$\pm$0.52} & 42.20{$\pm$2.91} & 95.77{$\pm$1.94} & 40.23{$\pm$4.30} & 92.98{$\pm$4.33} \\
& Restormer & 46.75{$\pm$0.76} & 98.65{$\pm$0.20} & 46.03{$\pm$0.81} & 98.46{$\pm$0.23} & 43.62{$\pm$2.08} & 97.51{$\pm$0.76} & 42.03{$\pm$3.38} & 96.37{$\pm$2.02} \\
& AMIR      & 46.92{$\pm$0.78} & 98.65{$\pm$0.21} & 46.20{$\pm$0.84} & 98.46{$\pm$0.24} & 43.82{$\pm$2.03} & 97.52{$\pm$0.89} & 42.39{$\pm$3.08} & 96.40{$\pm$1.95} \\
& CoreDiff  & 46.93{$\pm$0.77} & 98.63{$\pm$0.18} & 46.24{$\pm$0.81} & 98.46{$\pm$0.19} & 43.79{$\pm$2.03} & 97.52{$\pm$0.69} & 42.14{$\pm$3.25} & 96.21{$\pm$1.95} \\
& RDDM      & 47.06{$\pm$0.82} & 98.72{$\pm$0.21} & 46.36{$\pm$0.87} & 98.54{$\pm$0.24} & 43.59{$\pm$2.61} & 96.90{$\pm$1.95} & 41.59{$\pm$4.32} & 93.76{$\pm$6.23} \\
\rowcolor{cyan!20}\cellcolor{white}& \textbf{FoundDiff} & \textbf{47.37{$\pm$0.77}} & \textbf{98.84{$\pm$0.18}} & \textbf{46.72{$\pm$0.81}} & \textbf{98.70{$\pm$0.20}} & \textbf{44.42{$\pm$2.13}} & \textbf{97.87{$\pm$0.92}} & \textbf{42.64{$\pm$3.84}} & \textbf{96.51{$\pm$3.60}} \\
\hline

% --- Chest Region ---
\multirow{7}{*}{\textbf{Chest}}
& RED-CNN   & 35.68{$\pm$1.85} & 84.22{$\pm$5.06} & 35.42{$\pm$1.82} & 83.24{$\pm$5.26} & 34.72{$\pm$1.68} & 80.15{$\pm$5.55} & 34.33{$\pm$1.62} & 78.88{$\pm$5.28} \\
& PDF       & 35.58{$\pm$1.84} & 83.60{$\pm$5.00} & 35.32{$\pm$1.81} & 82.53{$\pm$5.06} & 34.40{$\pm$1.58} & 78.48{$\pm$5.00} & 34.07{$\pm$1.60} & 78.86{$\pm$5.09} \\
& Restormer & 35.78{$\pm$1.90} & 84.50{$\pm$5.44} & 35.54{$\pm$1.86} & 83.73{$\pm$5.60} & 34.92{$\pm$1.70} & 81.95{$\pm$5.83} & 34.57{$\pm$1.64} & 80.96{$\pm$5.94} \\
& AMIR      & 35.81{$\pm$1.88} & 84.46{$\pm$5.16} & 35.56{$\pm$1.86} & 83.71{$\pm$5.39} & 34.95{$\pm$1.71} & 82.06{$\pm$5.82} & 34.61{$\pm$1.65} & 81.08{$\pm$6.04} \\
& CoreDiff  & 35.73{$\pm$1.91} & 84.22{$\pm$5.52} & 35.49{$\pm$1.88} & 83.22{$\pm$5.59} & 34.85{$\pm$1.70} & 81.19{$\pm$5.49} & 34.43{$\pm$1.58} & 79.82{$\pm$5.19} \\
& RDDM      & 35.59{$\pm$1.85} & 84.35{$\pm$5.13} & 35.33{$\pm$1.82} & 83.52{$\pm$5.36} & 34.70{$\pm$1.65} & 81.64{$\pm$5.46} & 34.33{$\pm$1.63} & 80.59{$\pm$5.44} \\
\rowcolor{cyan!20}\cellcolor{white}& \textbf{FoundDiff} & \textbf{35.94{$\pm$1.86}} & \textbf{85.24{$\pm$4.94}} & \textbf{35.68{$\pm$1.85}} & \textbf{84.36{$\pm$5.34}} & \textbf{35.11{$\pm$1.77}} & \textbf{82.55{$\pm$5.91}} & \textbf{34.78{$\pm$1.74}} & \textbf{81.70{$\pm$6.00}} \\
\hline

% --- Head Region ---
\multirow{7}{*}{\textbf{Head}}
& RED-CNN   & 51.03{$\pm$6.23} & 99.26{$\pm$0.60} & 50.81{$\pm$6.13} & 99.25{$\pm$0.60} & 49.87{$\pm$5.70} & 99.17{$\pm$0.58} & 49.15{$\pm$5.42} & 99.05{$\pm$0.59} \\
& PDF       & 49.93{$\pm$5.77} & 99.16{$\pm$0.48} & 49.96{$\pm$5.94} & 99.14{$\pm$0.55} & 48.72{$\pm$5.52} & 98.92{$\pm$0.66} & 48.01{$\pm$5.24} & 98.71{$\pm$0.67} \\
& Restormer & 52.58{$\pm$6.15} & 99.56{$\pm$0.45} & 52.28{$\pm$6.06} & 99.54{$\pm$0.45} & 51.08{$\pm$5.70} & 99.44{$\pm$0.47} & 50.32{$\pm$5.50} & 99.36{$\pm$0.49} \\
& AMIR      & 52.55{$\pm$6.11} & 99.38{$\pm$0.53} & 52.30{$\pm$6.04} & 99.38{$\pm$0.52} & 51.18{$\pm$5.72} & 99.32{$\pm$0.50} & 50.45{$\pm$5.53} & 99.25{$\pm$0.51} \\
& CoreDiff  & 52.50{$\pm$6.15} & 99.53{$\pm$0.48} & 52.20{$\pm$6.06} & 99.50{$\pm$0.48} & 50.94{$\pm$5.69} & 99.36{$\pm$0.51} & 50.13{$\pm$5.47} & 99.24{$\pm$0.54} \\
& RDDM      & 53.78{$\pm$5.62} & 99.63{$\pm$0.39} & 53.45{$\pm$5.53} & 99.62{$\pm$0.40} & 52.08{$\pm$5.20} & 99.52{$\pm$0.42} & 51.23{$\pm$5.02} & 99.43{$\pm$0.44} \\
\rowcolor{cyan!20}\cellcolor{white}& \textbf{FoundDiff} & \textbf{54.21{$\pm$5.75}} & \textbf{99.69{$\pm$0.38}} & \textbf{53.86{$\pm$5.66}} & \textbf{99.67{$\pm$0.39}} & \textbf{52.48{$\pm$5.33}} & \textbf{99.58{$\pm$0.43}} & \textbf{51.66{$\pm$5.17}} & \textbf{99.51{$\pm$0.45}} \\
\shline
\end{tabular}}
\end{table*}

\begin{figure*}[!t]
\centerline{
\includegraphics[width=1\textwidth]{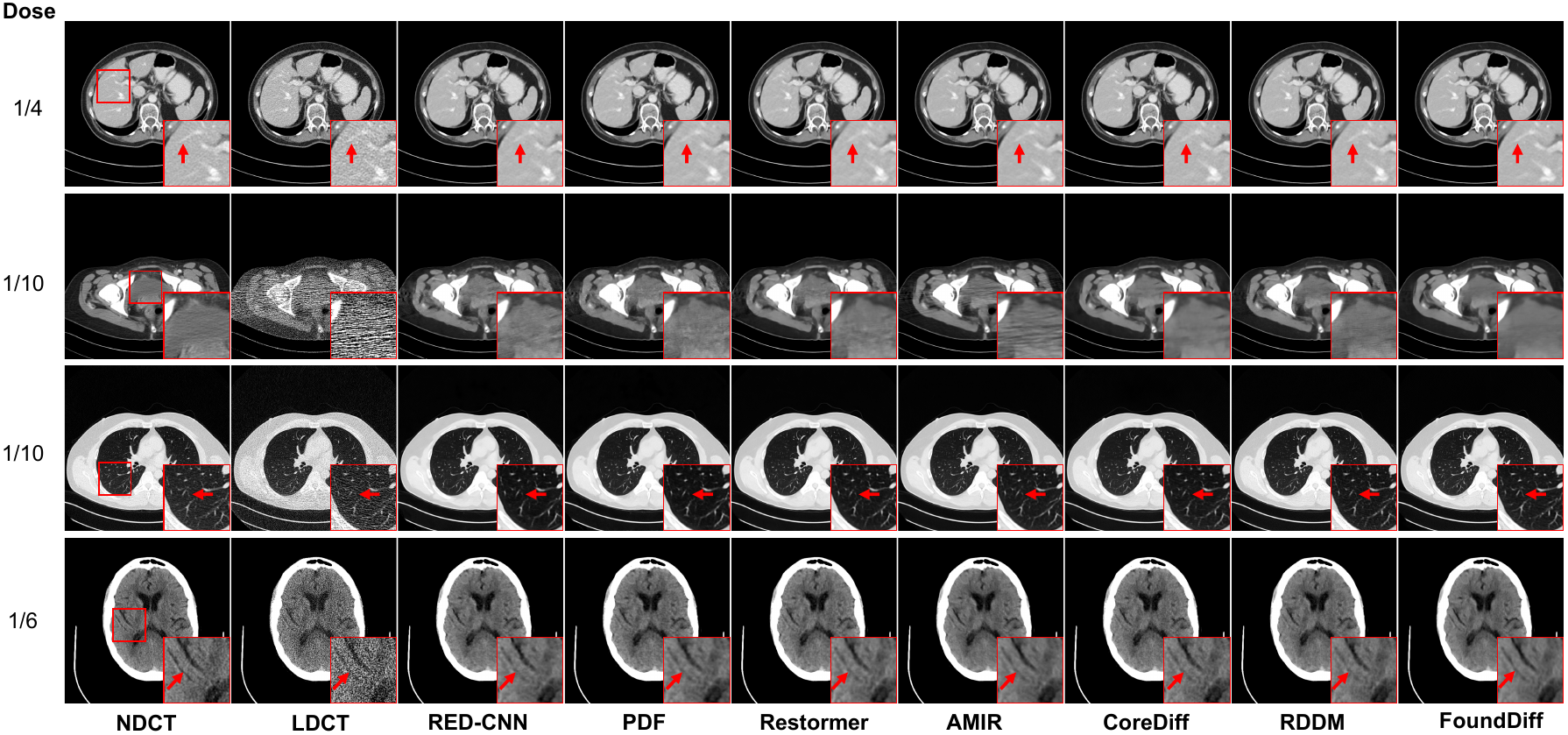}}
\caption{Qualitative results of CT images from the known dose levels. The red ROI is zoomed in for visual comparison. The first two rows show abdominal images with a display window of [-160, 240]~HU, the third row shows lung images with a display window of [-1350, 150]~HU, and the fourth row shows head images with a display window of [0, 80]~HU.}
\label{fig:known_dose}
\end{figure*}

\subsection{Evaluation of \modelname}
\label{eval_2020}
For training \modelname, we utilize the large simulated dataset. 
In perception stage, \stageone is trained on eight dose levels (1/2, 1/3, 1/4, 1/5, 1/6, 1/8, 1/10, and 1/20) to learn a continuous dose-aware representation. 
In denoising stage, \stagetwo is trained only on four dose levels (1/2, 1/4, 1/6, and 1/10). 
For a fair comparison, all competing methods are trained under the same four-dose setting used in \stagetwo. Since \stageone is frozen during denoising and only provides conditioning embeddings, the denoising comparison is conducted under the same training conditions.

For evaluation, we report denoising performance on the trained (seen) dose levels (1/2, 1/4, 1/6, 1/10) and unseen dose levels (1/5, 1/7, 1/15, 1/20) to assess generalization. 
In particular, 1/7 and 1/15 are strictly unseen, used by neither \stageone, \stagetwo, nor any competing method, enabling fair generalization assessment across components.
We further assess cross-dataset generalization on Mayo-2016, CQ500, and piglet datasets.

% For the training of \modelname, we utilize the large simulated dataset. 
% \stageone is trained on all data and \stagetwo is trained on four dose levels: 1/2, 1/4, 1/6, and 1/10. Testing of denoising performance is performed on both these trained dose levels and four additional unseen dose levels, 1/3, 1/5, 1/8, and 1/20, to evaluate the generalization ability of \modelname.
% Given that the primary focus of this work is unified LDCT denoising with robust generalization to new data without requiring additional fine-tuning, both \modelname and all competing methods are trained using the large simulated dataset under a multi-condition setting unless noted otherwise. 
%A separate comparison against models trained on individual conditions will be presented in the ablation studies section.

\begin{table}[!h]
\centering
\caption{Average perceptual metrics over seen and unseen dose levels in large simulated dataset.}
\label{tab:percep_metrics_avg_seen_unseen}
\setlength{\tabcolsep}{4pt}
\resizebox{\linewidth}{!}{
\begin{tabular}{lcccccccc}
\shline
\multirow{2}{*}{\textbf{Method}} &
\multicolumn{2}{c}{\textbf{LPIPS} $\downarrow$} &
\multicolumn{2}{c}{\textbf{VIF} $\uparrow$} &
\multicolumn{2}{c}{\textbf{GMSD} $\downarrow$} &
\multicolumn{2}{c}{\textbf{FSIM} $\uparrow$} \\
\cmidrule(lr){2-3}\cmidrule(lr){4-5}\cmidrule(lr){6-7}\cmidrule(lr){8-9}
& \textbf{Seen} & \textbf{Unseen} & \textbf{Seen} & \textbf{Unseen} & \textbf{Seen} & \textbf{Unseen} & \textbf{Seen} & \textbf{Unseen} \\
\hline
REDCNN    & 0.1454 & 0.1816 & 0.7006 & 0.6481 & 0.0141 & 0.0188 & 0.9898 & 0.9860 \\
PDF       & 0.1481 & 0.1905 & 0.6927 & 0.6367 & 0.0153 & 0.0204 & 0.9893 & 0.9847 \\
Restormer & 0.1490 & 0.1726 & 0.6962 & 0.6482 & 0.0133 & 0.0174 & 0.9899 & 0.9867 \\
AMIR      & 0.1455 & 0.1734 & 0.6984 & 0.6515 & 0.0131 & 0.0170 & 0.9903 & 0.9871 \\
Corediff  & 0.1272 & 0.1646 & 0.7040 & 0.6530 & 0.0129 & 0.0172 & 0.9904 & 0.9868 \\
RDDM      & \textbf{0.1179} & \textbf{0.1457} & 0.7055 & 0.6587 & 0.0123 & 0.0164 & 0.9912 & 0.9878 \\
\rowcolor{cyan!20} \textbf{FoundDiff} & 0.1340 & 0.1605 & \textbf{0.7134} & \textbf{0.6647} & \textbf{0.0121} & \textbf{0.0160} & \textbf{0.9914} & \textbf{0.9882} \\
\shline
\end{tabular}}
\end{table}

\subsubsection{Evaluation on seen dose levels}
Table~\ref{tab:seen_results} presents the quantitative testing results on the seen (training) dose levels. 
It can be observed that \modelname consistently outperforms competing methods across PSNR and SSIM on these known dose levels, indicating its superior denoising performance under different scanning conditions.
The baseline model RDDM achieves suboptimal
performance on head and abdominal data but yields less satisfactory results on chest data. 
Although AMIR includes a module designed for adaptive task recognition, its performance is still inferior to that of \modelname, which we specifically designed in \stageone to account for dose and anatomical information.
We note that PDF's performance falls short of expectations.
We attribute this primarily to its reliance on explicit dose level inputs, which do not adapt to the variations in noise patterns and structural textures across different anatomical regions.
 
Fig.~\ref{fig:known_dose} presents the qualitative results of representative denoised images on the known dose levels from three anatomical regions.
The red ROI is zoomed in for visual comparison and the key difference is pointed out by arrows.
In the first row, corresponding to a common 1/4 abdominal dose level, our method preserves clearer details, such as the white vessels visible. 
In the second row (1/10 dose), only our \modelname successfully suppresses noise without introducing artifacts, whereas other methods either leave some noise or blur the original tissue details. 
In the third row of 1/10 dose chest data, our \modelname restores details most closely matching those of the NDCT reference. 
Finally, in the last row of 1/6 head dose results, our method not only effectively suppresses noise but also preserves the structural details to the greatest extent.
The above results underscore the overwhelming effectiveness of our method for unified LDCT denoising across multiple anatomical regions and dose levels.

\begin{figure*}[!t]
\centerline{
\includegraphics[width=1\textwidth]{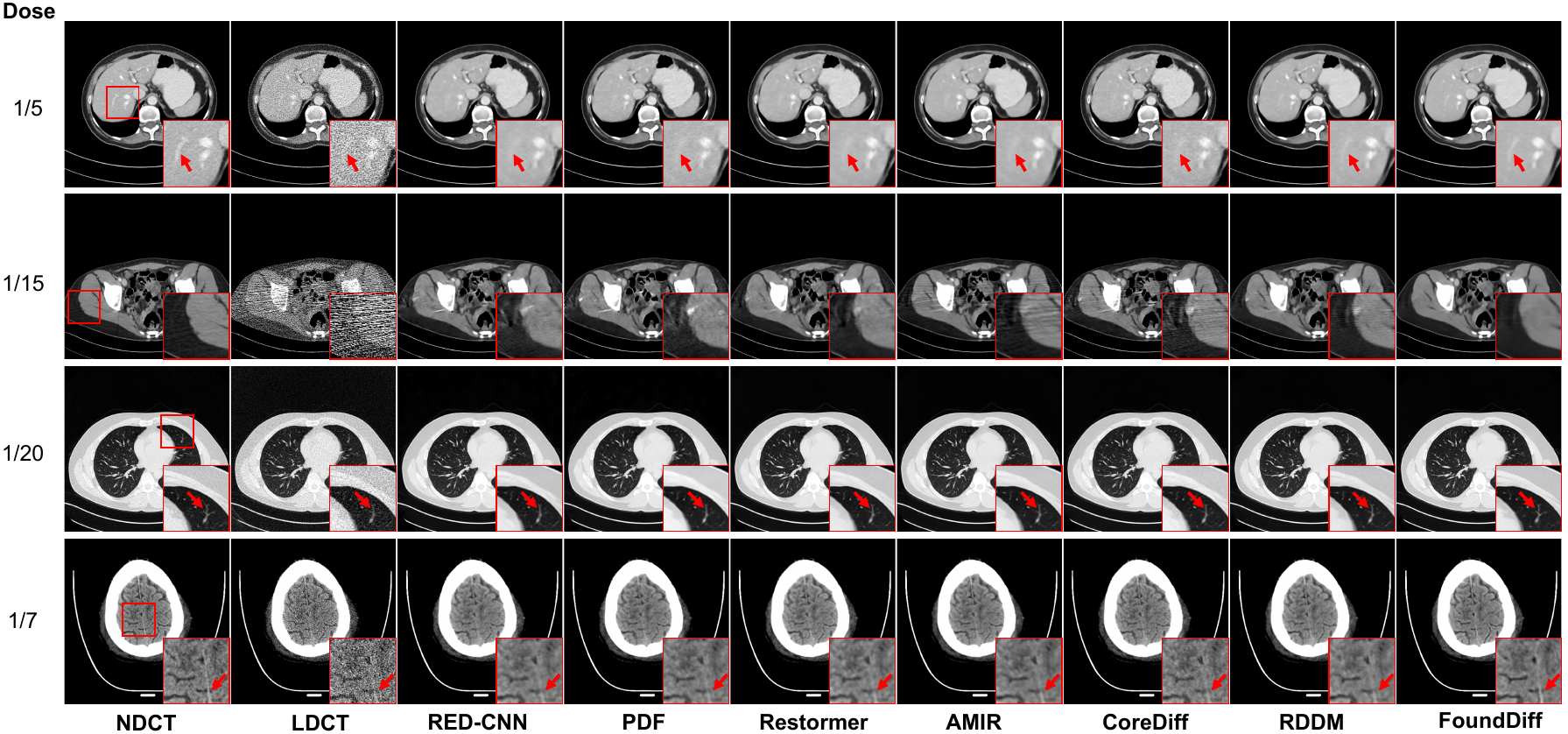}}
\caption{Qualitative results of CT images from the unseen dose levels. The red ROI is zoomed in for visual comparison. The first two rows show abdominal images with a display window of [-160, 240]~HU, the third row shows lung images with a display window of [-1350, 150]~HU, and the fourth row shows head images with a display window of [0, 80]~HU.}
\label{fig:unknown_dose}
\end{figure*}

\subsubsection{Evaluation on unseen dose levels}
To assess the generalization capability of \modelname beyond the trained dose levels, we further evaluate the model on four unseen dose levels (1/5, 1/7, 1/15, 1/20), where 1/7 and 1/15 dose levels are strictly unseen. 
As shown in Table~\ref{tab:unseen_results}, \modelname achieves superior quantitative results on unseen dose levels, significantly outperforming compared models across all evaluation metrics, which maintains robust denoising performance, especially in the ultra-low-dose level.
This highlights the effectiveness of the dose and anatomy representations learned by \stageone, particularly in modeling continuous ordinal relationships, enhancing the generalization to unseen dose levels.
As evidenced in qualitative examples in Fig.~\ref{fig:unknown_dose}, particularly in the challenging 1/20 ultra-low-dose cases (Rows 3), only our \modelname effectively restores integrity of anatomical structures. 
At the strictly unseen 1/15 dose level, competing methods tend to leave noise and introduce artifact-like structures, while \modelname more effectively suppresses noise with fewer artifacts; at the 1/7 dose level, \modelname better preserves fine anatomical details than the competing methods.
% Additionally, results for the dose levels shown in Rows 1 and 4 confirm that our \modelname produces finer details and textures compared to other methods.
These results highlight \modelname's robustness and its potential generalizability in diverse scanning protocols.

\begin{figure*}[!t]
\centerline{
\includegraphics[width=1\textwidth]{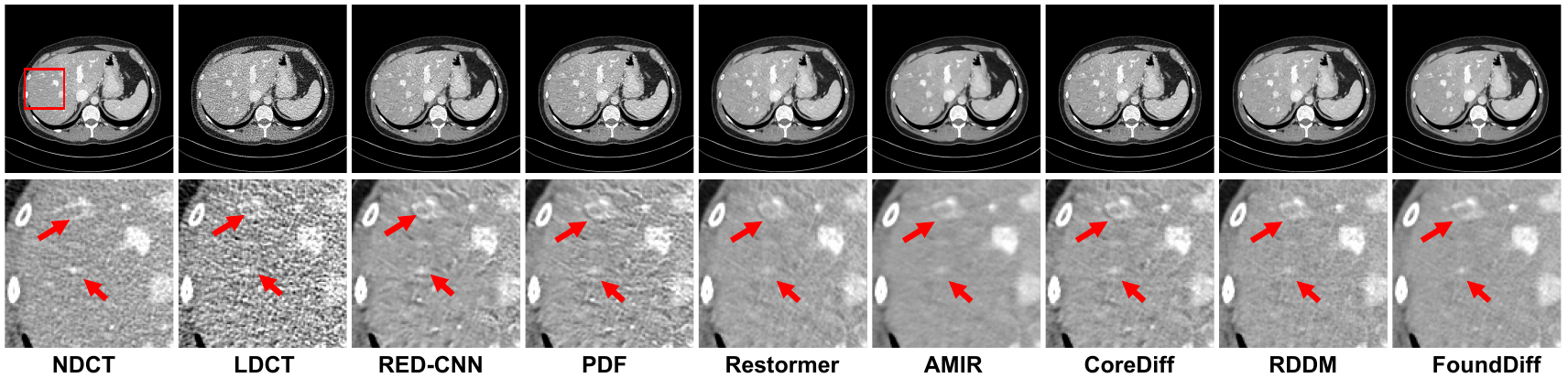}}
\caption{Qualitative results from unseen Mayo-2016 dataset. The red ROI is zoomed in for visual comparison. The display window is [-160, 240] HU.}
\label{fig:mayo_2016}
\end{figure*}

\begin{table*}[h]
\centering
\caption{Quantitative performance on the CQ500 dataset in terms of PSNR and  SSIM [$\times 10^{-2}$]. The best results are in \textbf{bold}.}
\label{tab:cq500_results}
\renewcommand\arraystretch{1}
\resizebox{\textwidth}{!}{
\begin{tabular}{l|cc|cc|cc|cc}
\shline
\multirow{2}{*}{\textbf{Method}}
& \multicolumn{2}{c|}{\textbf{1/4 Dose}}
& \multicolumn{2}{c|}{\textbf{1/7 Dose}}
& \multicolumn{2}{c|}{\textbf{1/10 Dose}}
& \multicolumn{2}{c}{\textbf{1/15 Dose}} \\
\cline{2-9}
& PSNR$\uparrow$ & SSIM$\uparrow$
& PSNR$\uparrow$ & SSIM$\uparrow$
& PSNR$\uparrow$ & SSIM$\uparrow$
& PSNR$\uparrow$ & SSIM$\uparrow$ \\
\hline
RED-CNN   & 55.52$\pm$3.66 & 99.79$\pm$0.21 & 54.92$\pm$3.45 & 99.76$\pm$0.26 & 54.45$\pm$3.32 & 99.75$\pm$0.27 & 53.82$\pm$3.17 & 99.73$\pm$0.26 \\
PDF       & 55.06$\pm$3.46 & 99.77$\pm$0.08 & 54.69$\pm$3.41 & 99.72$\pm$0.11 & 54.04$\pm$3.28 & 99.70$\pm$0.14 & 53.32$\pm$3.10 & 99.66$\pm$0.15 \\
Restormer & 56.25$\pm$2.94 & 99.84$\pm$0.07 & 55.48$\pm$2.67 & 99.80$\pm$0.07 & 54.90$\pm$2.54 & 99.77$\pm$0.07 & 54.20$\pm$2.45 & 99.72$\pm$0.08 \\
AMIR      & 55.94$\pm$2.87 & 99.72$\pm$0.10 & 55.28$\pm$2.72 & 99.71$\pm$0.10 & 54.74$\pm$2.63 & 99.67$\pm$0.09 & 54.09$\pm$2.55 & 99.61$\pm$0.10 \\
CoreDiff  & 53.65$\pm$2.17 & 99.75$\pm$0.05 & 53.29$\pm$2.08 & 99.67$\pm$0.07 & 52.96$\pm$2.03 & 99.61$\pm$0.08 & 52.40$\pm$1.99 & 99.52$\pm$0.10 \\
RDDM      & 56.40$\pm$2.98 & 99.84$\pm$0.27 & 55.83$\pm$2.41 & 99.77$\pm$0.24 & 55.21$\pm$3.47 & 99.67$\pm$0.23 & 54.25$\pm$3.08 & 99.66$\pm$0.28 \\
\rowcolor{cyan!20}\textbf{FoundDiff} & \textbf{56.88$\pm$3.19} & \textbf{99.88$\pm$0.05} & \textbf{56.06$\pm$2.90} & \textbf{99.84$\pm$0.07} & \textbf{55.39$\pm$2.69} & \textbf{99.79$\pm$0.09} & \textbf{54.55$\pm$2.47} & \textbf{99.75$\pm$0.12} \\
\shline
\end{tabular}}
\end{table*}

\begin{table*}[!h]
\centering
\caption{Quantitative performance on the piglet dataset in terms of PSNR and  SSIM [$\times 10^{-2}$]. The best results are in \textbf{bold}.}
\label{tab:piglet_doses1}
\renewcommand\arraystretch{1}
\resizebox{\textwidth}{!}{
\begin{tabular}{l|cc|cc|cc|cc}
\shline
\multirow{2}{*}{\textbf{Method}}
& \multicolumn{2}{c|}{\textbf{1/2 Dose}}
& \multicolumn{2}{c|}{\textbf{1/4 Dose}}
& \multicolumn{2}{c|}{\textbf{1/10 Dose}}
& \multicolumn{2}{c}{\textbf{1/20 Dose}} \\
\cline{2-9}
& PSNR$\uparrow$ & SSIM$\uparrow$
& PSNR$\uparrow$ & SSIM$\uparrow$
& PSNR$\uparrow$ & SSIM$\uparrow$
& PSNR$\uparrow$ & SSIM$\uparrow$ \\
\hline
RED-CNN & 42.78$\pm$2.37 & 98.14$\pm$0.48 & 40.15$\pm$3.56 & 97.41$\pm$0.95 & 40.57$\pm$2.33 & 97.52$\pm$0.90 & 37.89$\pm$2.55 & 95.47$\pm$1.70 \\

PDF & 42.42$\pm$2.33 & 97.73$\pm$0.46 & 40.49$\pm$3.83 & 97.60$\pm$0.99 & 40.38$\pm$2.41 & 97.38$\pm$0.98 & 37.70$\pm$2.60 & 95.30$\pm$1.70 \\

Restormer & 44.13$\pm$3.19 & 98.33$\pm$0.46 & 40.99$\pm$4.26 & 97.59$\pm$0.91 & 41.19$\pm$2.67 & 97.79$\pm$0.73 & 38.43$\pm$2.67 & 96.28$\pm$1.19 \\

AMIR & 43.63$\pm$2.63 & 98.37$\pm$0.43 & 40.62$\pm$3.84 & 97.65$\pm$0.89 & 41.08$\pm$2.34 & 97.80$\pm$0.66 & 38.36$\pm$2.54 & 96.33$\pm$1.16 \\

CoreDiff & 44.06$\pm$3.13 & 98.38$\pm$0.47 & 41.29$\pm$4.72 & 97.65$\pm$0.89 & 40.96$\pm$2.63 & 97.26$\pm$0.96 & 38.32$\pm$2.80 & 95.55$\pm$1.70 \\

RDDM & 44.20$\pm$3.25 & 98.40$\pm$0.46 & 41.04$\pm$4.32 & 97.70$\pm$0.90 & 41.14$\pm$2.68 & 97.80$\pm$0.92 & 38.38$\pm$2.63 & 96.08$\pm$1.07 \\
\rowcolor{cyan!20}\textbf{FoundDiff} & \textbf{44.60$\pm$3.55} & \textbf{98.50$\pm$0.46} & \textbf{41.58$\pm$2.82} & \textbf{98.13$\pm$0.64} & \textbf{41.26$\pm$4.56} & \textbf{97.88$\pm$0.90} & \textbf{38.59$\pm$2.75} & \textbf{96.57$\pm$1.04} \\
\shline
\end{tabular}}
\end{table*}

\subsubsection{Evaluation on structural and perceptual quality}
To complement PSNR and SSIM and provide a more comprehensive evaluation of denoising quality, we further assess the proposed method using additional perceptual and structure-aware metrics, including LPIPS, VIF, FSIM, and GMSD. 
As shown in Table~\ref{tab:percep_metrics_avg_seen_unseen}, averaged over all anatomical regions and dose levels, \modelname consistently achieves superior performance on VIF, FSIM, and GMSD, suggesting improved preservation of structural details under effective noise suppression; notably, recent studies have reported that VIF and FSIM correlate strongly with radiologists' diagnostic-quality scoring~\cite{mason2019comparison, sharma2025detail}.
Although its LPIPS scores are not optimal, this reflects a trade-off that prioritizes anatomical fidelity over perceptually sharp because it may reward noise-like patterns that resemble texture\cite{blau2018perception}.
Overall, these results indicate that \modelname achieves high fidelity under PSNR/SSIM while improving structural and perceptual quality.

\begin{table}[t]
\centering
\caption{Quantitative results (Mean$\pm$Std) on Mayo-2016 dataset.}
\label{table:2016}
\begin{tabularx}{\linewidth}{Xcc}
\shline
    & PSNR & SSIM [$\times 10^{-2}$] \\
 \hline
RED-CNN           &  43.58{$\pm$1.36}    &  96.82{$\pm$1.32}    \\

PDF   &  43.11{$\pm$1.75}    &  96.24{$\pm$1.81}   \\

Restormer    &  43.99{$\pm$1.39}    &  96.99{$\pm$1.16}    \\

AMIR  &  43.81{$\pm$1.51}    &  97.14{$\pm$1.04}  \\

CoreDiff  &  43.64{$\pm$1.39}    &   96.87{$\pm$1.34}   \\

RDDM   &   44.07$\pm$1.26 & 97.07$\pm$1.32  \\ 

\rowcolor{cyan!20} \modelname (\textbf{ours})  &\textbf{44.22$\pm$1.26}  & \textbf{97.31$\pm$1.24}   \\

\shline
\end{tabularx}
\end{table}

\subsubsection{Generalization to Mayo-2016 dataset}
\label{eval_2016}
To validate the adaptability of \modelname to new scenarios, we conduct cross-dataset generalization evaluation on the abdominal CT images with 1/4 dose from the Mayo-2016 dataset.
Table~\ref{table:2016} and Fig.~\ref{fig:mayo_2016} present the quantitative and representative qualitative results, respectively.
As reported in Table~\ref{table:2016}, \modelname demonstrates better performance on the Mayo-2016 dataset than other methods in terms of PSNR and SSIM.

Fig.~\ref{fig:mayo_2016} shows that images from the Mayo-2016 dataset exhibit significantly higher noise intensity compared to those from the simulated dataset. 
Consequently, methods such as RED-CNN and PDF, which already demonstrated suboptimal performance at ultra-low-dose levels on the simulated data, exhibit a marked degradation in generalization performance on the Mayo-2016 dataset.
Although Restormer, AMIR, and CoreDiff effectively suppress noise, Restormer and AMIR lose important structural details, while CoreDiff introduces artifacts.
The second-best method, RDDM, tends to compromise fine anatomical textures and may introduce artifacts.
In contrast, \modelname\ effectively suppresses noise while preserving critical anatomical details~(\eg, vessels), as indicated by the red arrow.
This demonstrates that our adaptive and precise recognition of dose level and anatomical region can effectively generalize to clinical LDCT data, thereby enhancing its clinical applicability and paving the way for universal CT imaging in real-world scenarios.

\subsubsection{Generalization to CQ500 dataset}
\begin{figure*}[!t]
\centerline{
\includegraphics[width=1\textwidth]{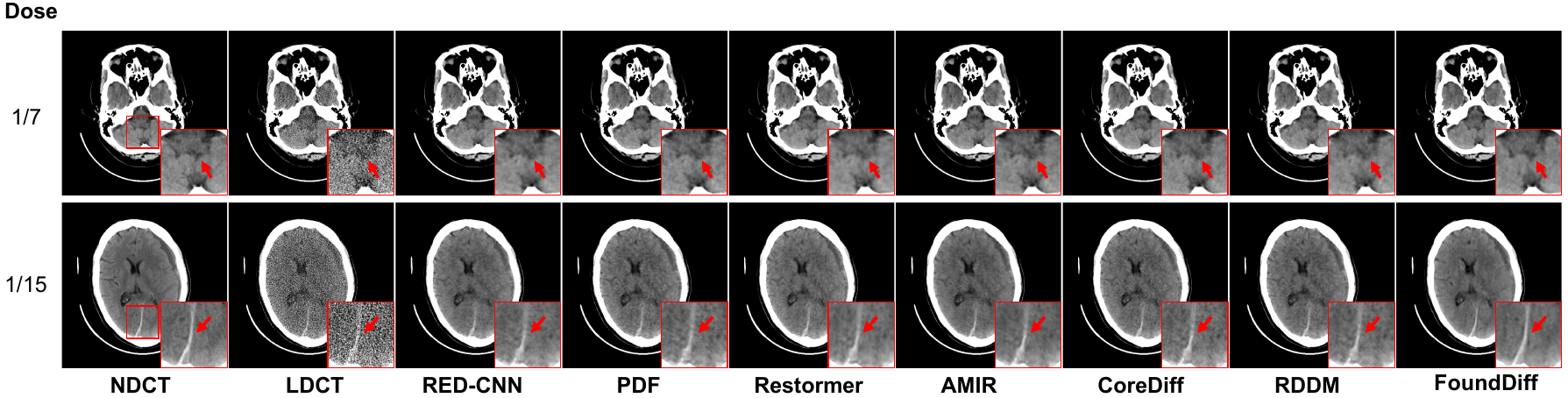}}
\caption{Qualitative results of some representative cases in CQ500 dataset. The red ROI is zoomed in for visual comparison.}
\label{fig:cq500_results}
\end{figure*}

To evaluate generalization across different scanners and institutions, we conduct testing experiments on CQ500 dataset. 
The quantitative results are summarized in Table~\ref{tab:cq500_results}.
Our \modelname demonstrates robust performance across all dose levels, even for strictly unseen 1/7 and 1/15 dose levels, indicating strong generalization to new data distributions.
In the representative examples of completely unseen dose levels, shown in Fig.~\ref{fig:cq500_results}, it can be observed that \modelname retains anatomical details more clearly than comparison methods, while effectively suppressing noise.
These results suggest that \modelname can generalize to multi-center scanner/institution data while preserving fine anatomical structures.

\begin{figure*}[!t]
\centerline{
\includegraphics[width=1\textwidth]{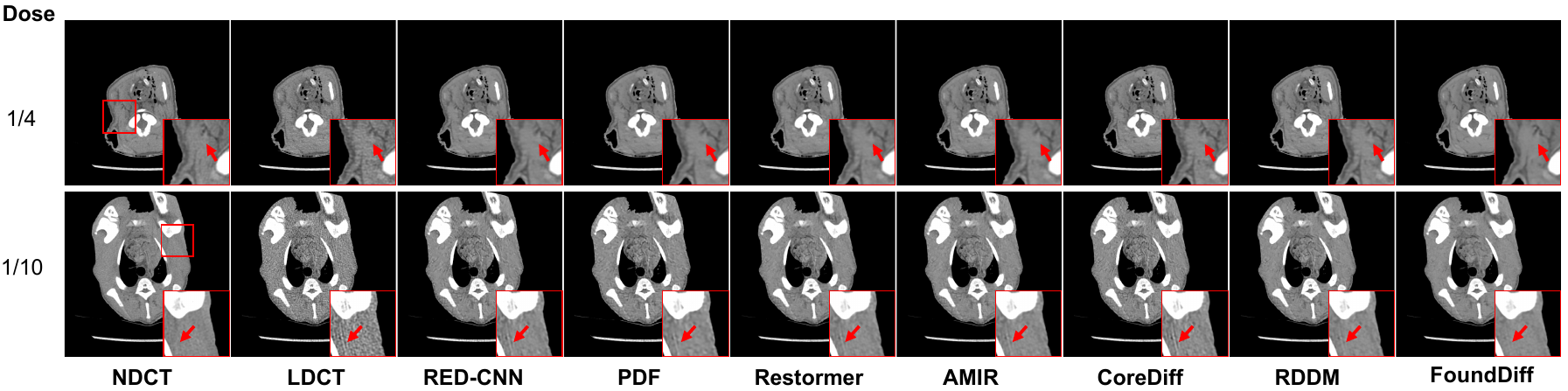}}
\caption{Qualitative results of some representative cases in piglet dataset. The red ROI is zoomed in for visual comparison.}
\label{fig:piglet_results}
\end{figure*}

\subsubsection{Generalization to piglet dataset}
To assess cross-anatomical generalization on real LDCT scans, we further evaluate \modelname on the piglet dataset.
Table~\ref{tab:piglet_doses1} demonstrates that \modelname achieves the best quantitative performance across all evaluation metrics.
Fig.~\ref{fig:piglet_results} 
presents two representative examples of denoised images. 
The first row shows that \modelname effectively suppresses noise while preserving fine anatomical details, outperforming all competing methods. 
As indicated by the arrows in the second row, artifacts arising in LDCT images are not adequately removed by the comparison methods, whereas our method successfully eliminates these artifacts. 
This demonstrates the strong cross-anatomy generalization capability of our approach on real LDCT scans.

\begin{table}[h]
\centering
\renewcommand\arraystretch{1}
\caption{Training detail of the \modelname and other methods in the large simulated dataset.}
\label{tab:train_detail}
\resizebox{\linewidth}{!}{
\begin{tabular}{lrrrr}
\shline
Method    & Parms. [M] & Batch size& Time/epoch [h] & Inference [s/image]\\
 \hline
RED-CNN    & 1.8  &4 & 0.88  &   0.038  \\

PDF    & 2.0  &4 & 1.22  &   0.045  \\

Restormer  & 13.0 & 4 &1.04&  0.102   \\

AMIR  &  12.7 & 4 &  2.17 &  0.124   \\

CoreDiff  & 6.5 &4 & 1.36 &   0.169  \\

RDDM  & 18.9 & 2 & 3.18 &   0.238  \\

\modelname  & 21.6 & 2 & 3.95  &  0.327 \\

\shline
\end{tabular}}
\end{table}

\subsubsection{Computational efficiency analysis}
Table~\ref{tab:train_detail} summarizes the computational cost of all methods.
To ensure a fair comparison given the varying memory footprints and batch sizes, we report the average time per epoch (calculated as equivalent epochs for iteration-based methods). 
\modelname remains computationally competitive among diffusion-based methods while improving denoising quality. 
Compared with RDDM, \modelname has a similar parameter scale with only slightly higher training cost (3.95 vs. 3.18 hours/epoch) and sub-second inference latency.
Compared with CoreDiff, despite a larger model (21.6M vs. 6.5M), \modelname achieves comparable inference time (0.327 vs. 0.169 s/image) due to efficient two-step sampling.
Overall, our method incurs only moderate overhead and offers a favorable efficiency-performance trade-off.

\subsubsection{Comparison with SOTA specialized models}
While the primary focus of our work is to achieve unified denoising across diverse conditions, we also benchmark \modelname against the SOTA specialized models trained on individual condition. 
% This comparison is crucial as it demonstrates whether our unified approach can maintain competitive performance in specialized scenarios without sacrificing the versatility that allows it to handle multiple conditions simultaneously.
To this end, we conduct a comparative analysis against two baseline methods, RDDM and Restormer, each trained specifically on three distinct conditions: 1/10 dose abdomen, 1/10 dose chest, and 1/10 dose head.
We select the 1/10 dose level because it represents a challenging scenario in LDCT imaging, making it more meaningful for single-dose performance comparisons.
Note that our \modelname maintains the unified training conditions.

Table~\ref{table:single} presents the quantitative results.
% As observed, while RDDM individually trained for the 1/4 dose abdomen subset achieves a higher PSNR, its performance significantly degrades when trained and tested on the 1/10 dose chest subset. 
\modelname effectively learns to integrate and exploit informative features from diverse inputs, enabling it to maintain optimal performance even compared to specialized models.
% Especially in data-scarce scenarios, such as the head region, our \modelname achieves superior performance, benefiting from its exposure to more diverse training data.
%Such evaluation provides insights into potential trade-offs between generalization and specialization, validating that our model remains effective even when compared to purpose-built alternatives.
Furthermore, it is crucial to emphasize that such individually trained models inherently lack the ability to generalize to other dose levels or anatomical regions, nor can they effectively process previously unseen dose conditions.

\begin{table}[h]
\centering
\caption{Quantitative results~(Mean$\pm$Std) comparing \modelname with specialized models.}
\label{table:single}
\begin{tabularx}{\linewidth}{Xlcc}
\shline
  Target task & Method  &PSNR& SSIM [$\times 10^{-2}$] \\

 \hline

 \multirow{3}*{1/10 dose abdomen} & Restormer    &  45.51{$\pm$0.95}    &  98.33{$\pm$0.27}     \\

& RDDM   &   45.61$\pm$1.05 & 98.36$\pm$0.30    \\ 

\rowcolor{cyan!20}\cellcolor{white}& \modelname &\textbf{45.94$\pm$1.09}  & \textbf{98.49$\pm$0.29}  \\

 \hline

 \multirow{3}*{1/10 dose chest} & Restormer    &  34.27{$\pm$2.77}    &  82.12{$\pm$6.22}     \\

& RDDM   &   33.68$\pm$2.76 & 80.43$\pm$6.55    \\ 

\rowcolor{cyan!20}\cellcolor{white}& \modelname &\textbf{34.39$\pm$2.79}  & \textbf{82.29$\pm$6.31}  \\

 \midrule

 \multirow{3}*{1/10 dose head} & Restormer    &  51.63{$\pm$5.17}    &  99.48{$\pm$0.49}     \\

& RDDM   &   52.57$\pm$5.33 & 99.54{$\pm$0.48}   \\ 

\rowcolor{cyan!20}\cellcolor{white}& \modelname &\textbf{52.93$\pm$4.89}  & \textbf{99.61$\pm$0.46}  \\

\shline
\end{tabularx}
\end{table}

\begin{table}[t]
\caption{Ablation study of \stageone on the dose prediction in terms of PLCC and SROCC.}
\label{tab:dose_prediction}
\centering
\begin{tabularx}{\linewidth}{Xcc}
\shline
 & PLCC  & SROCC  \\
\hline
CLIPIQA & 0.2691 & 0.2539  \\
CLIPIQA+ & 0.8617 &0.8792  \\
\stageone w/o $\mathcal{L}_\mathrm{rank}$ & 0.9337 & 0.8929\\
\stageone w/o $\mathcal{L}_\mathrm{a}$ &  0.9426 &0.9447
\\
\stageone (replace $\mathcal{L}_\mathrm{a}$ with $\mathcal{L}_\mathrm{cls}$) &  0.9522 & 0.9498 \\
\rowcolor{cyan!20}\stageone  & \textbf{0.9907} & \textbf{0.9831} \\
\shline
\end{tabularx}
\end{table}

% \begin{table}[h] 
% \centering 
% \caption{Average quantitative comparison among \modelname-CLS, CLIP-Diff, and \modelname.\shan{can be merged into other table?}} 
% \label{tab:foundDiff-Cls_vs_founddiff} 
% \begin{tabular*}{1\linewidth}{@{\extracolsep{\fill}}l|cc|cc}
% % \begin{tabular}{l|cc|cc} 
% \shline
% \multirow{2}{*}{} & \multicolumn{2}{c|}{\textbf{Seen Doses (Avg.)}} & \multicolumn{2}{c}{\textbf{Unseen Doses (Avg.)}} \\
% & PSNR & SSIM[$\times 10^{-2}$] & PSNR & SSIM[$\times 10^{-2}$] \\
% \hline 
% \modelname-CLS& 41.24 & 90.89 & 39.09 & 89.85 \\
% CLIP-Diff & 40.72 & 90.78 & 38.45 & 88.75 \\
% \textbf{\modelname (Ours)} & \textbf{41.75} & \textbf{91.08} & \textbf{41.08} & \textbf{90.27} \\ 
% \shline
% \end{tabular*} 
% \end{table}

\begin{figure}[t]
\centering
\includegraphics[width=1\linewidth]{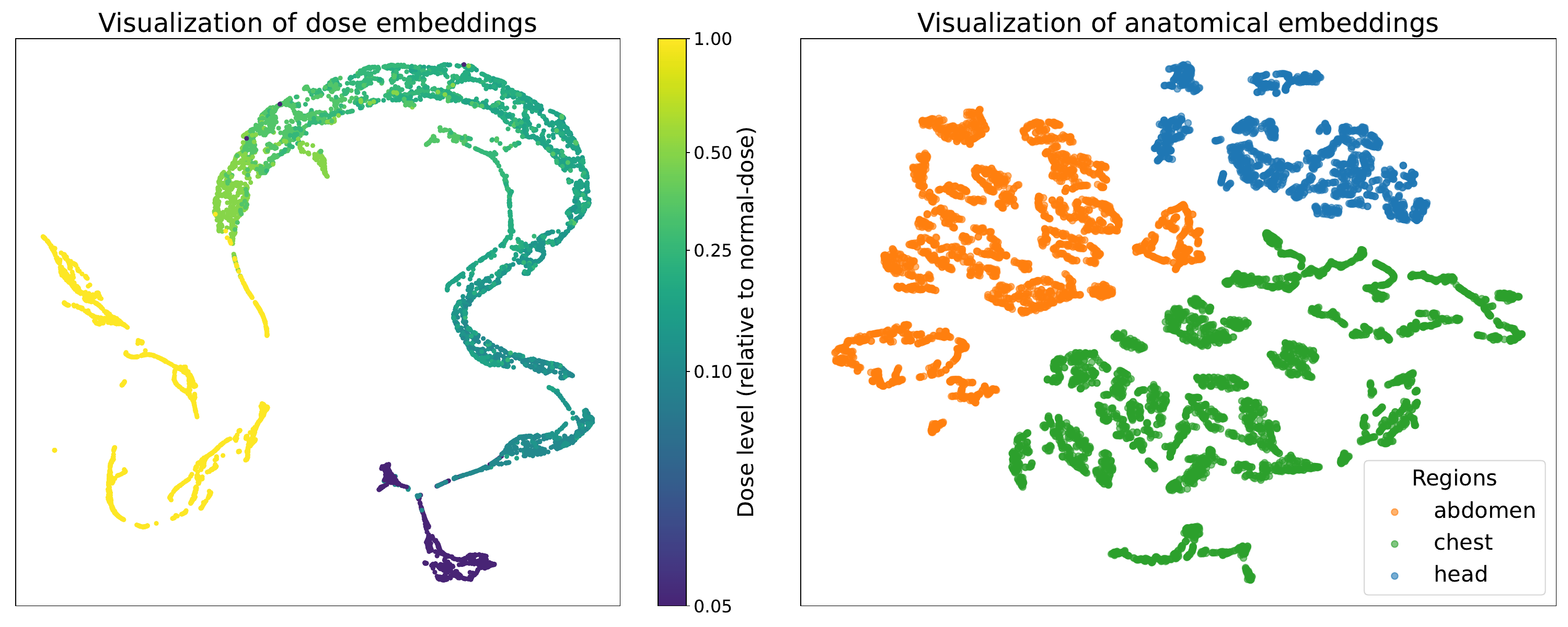}
\caption{t-SNE visualization of dose embeddings (left) and anatomical embeddings (right) generated from \stageone.}
\label{fig:feat_visual}
\end{figure}

\subsection{Ablation Studies}
\label{sec:ablation}
\subsubsection{Ablation on \stageone}
To validate the effectiveness of \stageone, we conduct the following ablation experiments.

\textbf{(i) Ablation on $\mathcal{L}_\mathrm{rank}$ and $\mathcal{L}_\mathrm{a}$.}
Table~\ref{tab:dose_prediction} indicates \stageone outperforms  the baseline CLIPIQA~\cite{wang2023exploring} and its fine-tuning version CLIPIQA+ in terms of both PLCC and SROCC. 
When the ranking loss ($\mathcal{L}_\mathrm{rank}$) is omitted, the ablated versions of \stageone exhibit performance degradation. 
This emphasizes the pivotal role that capturing the ordinal relationships of dose plays in dose prediction.
Furthermore, the results demonstrate that the anatomy discrimination loss ($\mathcal{L}_\mathrm{a}$) facilitates more accurate dose level prediction, which further highlights the importance of addressing anatomical heterogeneity in CT IQA.

In addition, we replace the contrastive loss $\mathcal{L}_\mathrm{a}$ with a cross-entropy classification objective, denoted as $\mathcal{L}_\mathrm{cls}$, to examine whether supervised contrastive learning is necessary compared with standard classification. 
We refer to the \stagetwo variant trained with $\mathcal{L}_\mathrm{cls}$ as \modelname-CLS.
Tables~\ref{tab:dose_prediction} and~\ref{table:ablation1} show that \modelname-CLS yields a significant reduction in dose prediction performance and decreases the denoising PSNR.
Consistent with prior studies~\cite{islam2021broad, graf2021dissecting}, these results support the necessity of contrastive learning for producing better-structured embeddings with improved clustering and larger decision margins than cross-entropy training.

\textbf{(ii) Feature visualization.} 
Fig.~\ref{fig:feat_visual} provides a clear visualization of the embeddings generated by \stageone. 
%The dose embeddings (left) are mapped according to their relative dose levels, with the color bar showing the percentage relative to normal-dose. 
\stageone well organizes dose-related information in a continuous manner, effectively differentiating dose features concentrated between 10\% and 50\%. 
The anatomical embeddings (right) further highlight the model's ability to separate different anatomical regions, indicating that the model successfully retains important anatomical information while predicting dose levels.

\textbf{(iii) Comparison with explicit metadata conditioning.}
To evaluate the necessity of \stageone, we implement two baselines using a frozen CLIP text encoder~\cite{clip} with metadata prompts: ``Low-dose CT image of the \{\texttt{anatomy}\} at \{\texttt{dose}\} dose level''. In CLIP-Diff, the text embeddings are injected into the diffusion model via DACB. In CLIP-Restormer, they are injected into Restormer through cross-attention, serving as a strong feed-forward baseline. Table~\ref{table:ablation1} shows that \modelname outperforms both baselines, indicating that the continuous ordinal dose representation and adaptive anatomical embeddings learned by \stageone are more effective than discrete metadata prompts, regardless of the backbone architecture.

\begin{table}[h]
\centering
\caption{Ablation study of \modelname components on seen dose levels (average metrics across all test images).}
\label{table:ablation1}
% \begin{tabular*}{1\linewidth}{@{\extracolsep{\fill}}lcc}
\begin{tabularx}{\linewidth}{Xcc}
\shline 
 &PSNR& SSIM [$\times 10^{-2}$]  \\ 
  \hline
 UNet (in RDDM) & 41.20 & 90.31  \\
 UNet (in RDDM)+DACB & 41.31 & 90.44 \\
 \hline
 CLIP-Restormer & 41.27 & 90.50 \\
 CLIP-Diff & 41.32 & 90.58  \\
 \modelname-CLS& 41.43 & 90.80 \\
 \hline
RDDM  
&  41.29    &  90.38   
 \\

\quad + Dose condition   
&  41.44    &  90.74  
  \\

\quad + Anatomical condition (cross-attention)
&  41.39    &  90.67  
  \\
  
\quad + Anatomical condition (CSSM)  
&  41.50    &  90.77  

  \\

\quad\quad  + Transposed attention  
&  41.56    &  90.82   
  \\

\rowcolor{cyan!20}\quad\quad\quad  + Dose condition (\textbf{\modelname})    
&  \textbf{41.75}    &  \textbf{91.08}   
  \\

\shline
\end{tabularx}
% \end{tabular*}
\end{table}

\subsubsection{Ablation on \stagetwo}
To validate the effectiveness of the components in \stagetwo during denoising, we conduct experiments from the baseline RDDM equipped only with the RLEB and integrate the core components of \stagetwo. 
Table~\ref{table:ablation1} shows incremental performance gains with the addition of each component. 
Compared to the anatomical integration of cross-attention~\cite{luo2023controlling}, which incurs quadratic complexity, our CSSM yields better performance with only linear complexity.
Notably, the integration of the dose and anatomical conditioning mechanisms yields substantial improvements, underscoring the critical importance of jointly exploring noise characteristics and anatomical heterogeneity during the denoising process.

Furthermore, to isolate the effect of diffusion, we remove the diffusion formulation and use RDDM's U-Net as a direct feed-forward baseline. Comparing the U-Net baseline with full RDDM shows that diffusion itself provides a measurable gain. In addition, DACB brings a +0.11 dB improvement in the feed-forward setting, but a larger +0.46 dB gain in the diffusion framework, validating the synergy between DACB's dose-timestep fusion and the residual diffusion process.

%% file: sections/discussion.tex
\section{Discussion}
\label{sec:discussion}
\subsection{Novelty of \modelname}
%While our framework leverages foundational architectures (\eg, CLIP and Mamba), it goes beyond a direct combination by substantially redesigning their internal mechanisms to accommodate the unique characteristics of LDCT imaging.
At the task level, we formulate a unified LDCT denoising setting that targets generalization across heterogeneous dose levels and anatomical regions, addressing the limitation of prior single-setting training.
Methodologically, we introduce a two-stage design: \stageone learns continuous, semantically structured dose/anatomy embeddings via ranking and anatomy-discriminative objectives, providing an adaptive prior. 
\stagetwo injects these priors through a tailored conditioning block: global dose modulation and anatomy-conditioned linear-complexity scanning, enabling adaptive denoising without substantial overhead.

\subsection{Benefits of \modelname}
\modelname achieves superior unified and generalizable LDCT denoising performance by enhancing adaptive dose- and anatomical-aware capabilities. 
We discuss \modelname's key advantages over competing methods as follows.
{(\textbf{i})} We highlight the crucial role of \stageone in jointly learning dose-related and anatomy-specific representations within a unified architecture. Unlike previous IQA-based methods such as CLIPIQA~\cite{wang2023exploring}, which rely on MSE loss for direct quality regression, \stageone introduces a dose-ranking loss that captures the ordinal structure of dose levels, enabling improved generalization to unseen doses through continuous feature learning.
Furthermore, our anatomy discrimination loss effectively distinguishes anatomical regions, facilitating the extension of \modelname to additional body parts, such as the neck and joints.
{(\textbf{ii})} Building upon the efficient SOTA baseline (RDDM), \modelname incorporates two specially designed conditioning strategies within the proposed \block, which implicitly inject both the dose and anatomy features into the denoising process.
This design enables unified and adaptive LDCT denoising across a wide range of dose levels and anatomical regions. 
Notably, this adaptability generalizes well to unseen dose levels without requiring explicit condition inputs or additional fine-tuning.
{(\textbf{iii})} \modelname operates entirely in the image domain without requiring access to raw projection data or explicit metadata, making it vendor-agnostic and compatible with standard reconstructed CT images for plug-and-play deployment. Furthermore, since \stageone directly perceives image-level quality, it naturally captures different degradation characteristics when image appearance varies due to factors such as reconstruction kernels, even under the same nominal dose level.
%In contrast to raw-data-based approaches that are tightly coupled to specific hardware or reconstruction algorithms~\cite{need,wang2020deep}, our design significantly enhances clinical applicability and scalability.

Furthermore, we highlight the significant ability of recovering reliable structural integrity in complex anatomical regions  under ultra-
low-dose conditions. 
As shown in Figs.~\ref{fig:known_dose} and~\ref{fig:unknown_dose}, at relatively higher dose levels (\eg, 1/2 and 1/4), \modelname offers limited visual advantage since most SOTA methods already produce satisfactory results.
In contrast, for the two rows (seen 1/10 and strictly unseen 1/15 dose abdominal scans), where anatomical structures are severely corrupted by noise, the performance gap becomes pronounced. 
Under such extreme degradation, compared methods often leave residual noise and may introduce structural artifacts or distortions, whereas \modelname achieves stronger denoising while better preserving anatomy.

\subsection{Limitation and Future Work}
Here, we acknowledge some limitations.
First, while our simulation aims for clinical realism, discrepancies persist between the simulated data and the heterogeneous nature of real-world clinical LDCT scans.
Second, although we introduce efficient SSM and transposed attention within our block design and reduce the inference sampling steps to 2, full-image processing and diffusion-based denoising still entail higher computational costs compared to traditional models. 
Third, in the challenging ultra-low-dose setting, the restored results may appear mildly over-smoothing; nevertheless, our method is the only one that completely suppresses the heavy noise without introducing structural artifacts, indicating a trade-off toward artifact-free reconstruction.

Looking forward, several avenues warrant further investigation. 
First, enhancing real-world generalization remains crucial; this involves acquiring more diverse clinical datasets through institutional collaborations and robust domain adaptation techniques to bridge the simulation-to-reality gap. 
Second, exploring more efficient generative denoising frameworks, such as autoregressive models~\cite{var}, could address current inference latency and cost limitations. 
Third, incorporating language model-guided semantic information~\cite{chen2025langmamba} or  distillation with high dose model~\cite{li2024quad} during training may help preserve finer details under ultra-low-dose settings and enhance generalization.
Furthermore, extending \stageone's perception capability to incorporate broader CT imaging conditions such as reconstruction kernels and scanner vendors could evolve \modelname from dose/anatomy-aware conditioning toward imaging-aware conditioning for clinical deployment.

%% file: sections/conclusion.tex
\section{Conclusion}
\label{sec:conclusion}

In this work, we introduced FoundDiff, a two-stage foundational diffusion model designed for unified and adaptive LDCT denoising across varied dose levels and anatomical regions. 
FoundDiff incorporates  a \stageone in the first stage, trained with dose-ranking and anatomy discrimination losses to extract continuous dose-aware and anatomy-specific semantic embeddings,  and  a \stagetwo in the second stage that integrates these embeddings through the proposed \block, enabling adaptive denoising without explicit condition input.  
Extensive experiments validate the effectiveness, adaptability, and generalizability of \modelname under diverse scanning settings, highlighting its potential as a versatile and clinically deployable solution for LDCT denoising.

% First, the \stageone, trained via dose ranking and anatomy discrimination losses, extracts ordinal dose and anatomical semantic embeddings. 
% Second, operating within a residual denoising diffusion framework, the \stagetwo model utilizes these embeddings via its core \block block, which employs timestep-fused adaptive normalization and CSSM, for effective, adaptive denoising.
% Extensive experiments confirm that \modelname significantly outperforms compared methods, demonstrating superior performance and remarkable generalization across varied conditions. \modelname signifies a substantial advancement towards versatile and clinically applicable LDCT denoising solutions.